\documentclass[10pt,twocolumn,letterpaper]{article}

\usepackage[pagenumbers]{cvpr} % To force page numbers, e.g. for an arXiv version

\makeatletter
\@namedef{ver@everyshi.sty}{}
\makeatother
\usepackage{tikz}

\usepackage{graphicx}
\usepackage{tikz}
\usepackage{comment}
\usepackage{amsmath,amssymb} % define this before the line numbering.
\usepackage{color}

\usepackage{graphicx}
\usepackage{caption}
\usepackage{subcaption}

\usepackage{times}
\usepackage{epsfig}
\usepackage{graphicx}
\usepackage{amsmath}
\usepackage{amssymb}
\usepackage{dsfont}

\usepackage{url}       
\usepackage{booktabs}   
\usepackage{amsfonts}  
\usepackage{nicefrac}
\usepackage{microtype} 
\usepackage{graphicx}
\usepackage{caption}
\usepackage{comment}
\usepackage{multirow}
\usepackage{multicol}
\usepackage{stfloats}

\usepackage[pagebackref,breaklinks,colorlinks]{hyperref}

\usepackage[capitalize]{cleveref}
\crefname{section}{Sec.}{Secs.}
\Crefname{section}{Section}{Sections}
\Crefname{table}{Table}{Tables}
\crefname{table}{Tab.}{Tabs.}

\def\cvprPaperID{6679} % *** Enter the CVPR Paper ID here
\def\confName{CVPR}
\def\confYear{2023}

\usepackage{enumitem} %< control spacing in itemize/enumerate/...
\usepackage{overpic} %< add raw math symbols to figures
\usepackage{color}
\definecolor{turquoise}{cmyk}{0.65,0,0.1,0.3}
\definecolor{purple}{rgb}{0.65,0,0.65}
\definecolor{dark_green}{rgb}{0, 0.5, 0}
\definecolor{orange}{rgb}{0.8, 0.6, 0.2}
\definecolor{red}{rgb}{0.8, 0.2, 0.2}
\definecolor{darkred}{rgb}{0.6, 0.1, 0.05}
\definecolor{blueish}{rgb}{0.0, 0.3, .6}
\definecolor{light_gray}{rgb}{0.7, 0.7, .7}
\definecolor{pink}{rgb}{1, 0, 1}
\definecolor{greyblue}{rgb}{0.25, 0.25, 1}

\usepackage{blindtext}

\renewcommand{\paragraph}[1]{\vspace{1em}\noindent\textbf{#1}.}
\begin{document}

\newcommand{\OURS}{NPPs}
\newcommand{\OURSFULL}{Neural Part Priors}

%\title{[CVPR2022] Official LaTeX Template}
\title{\OURSFULL{}: Learning to Optimize Part-Based Object Completion in RGB-D Scans} %TODO update
% optimized part priors

\author{Andrea Tagliasacchi\\
Google Research \& University of Toronto\\
{\tt\small taglia@google.com}
% \and
}

\setlength{\abovedisplayskip}{2pt}
\setlength{\belowdisplayskip}{2pt}

\title{\OURSFULL{}: Learning to Optimize Part-Based Object Completion in RGB-D Scans}  % **** Enter the paper title here

\author{Alexey Bokhovkin\\
Technical University of Munich\\
{\tt\small aleksei.bokhovkin@tum.de}
\and
Angela Dai\\
Technical University of Munich\\
{\tt\small angela.dai@tum.de}
}

%%%%%%%%% TEASER
 \twocolumn[{%
 	\renewcommand\twocolumn[1][]{#1}%
 	\maketitle
 	\begin{center}
 		\vspace{-0.8cm}
 		\includegraphics[width=0.9\linewidth]{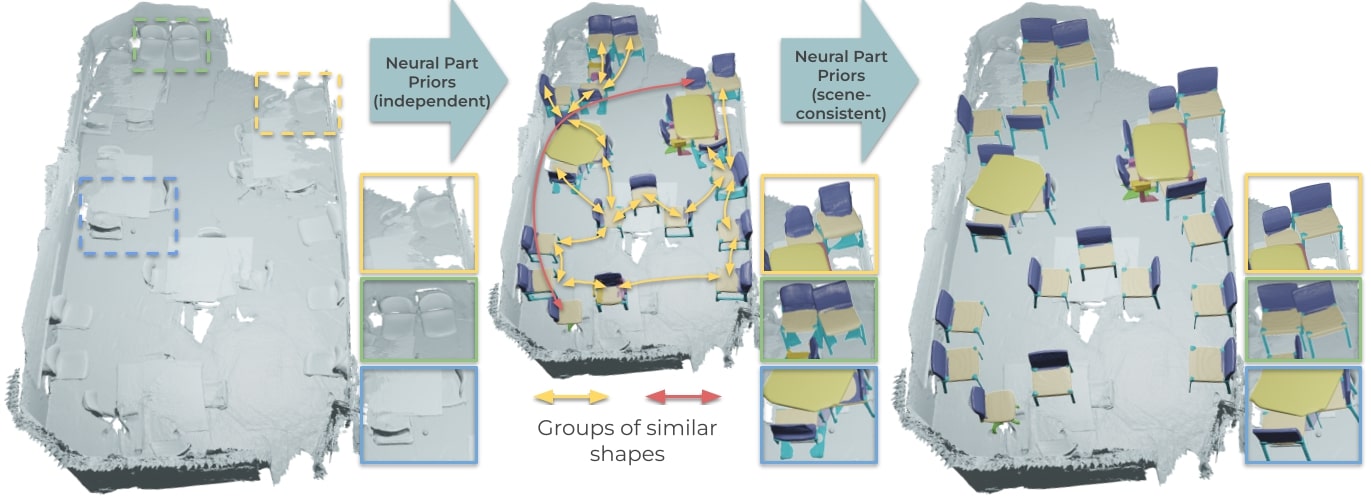}
 		\vspace{-0.2cm}
 		\captionof{figure}{
             Our \OURSFULL{} learn optimizable parametric latent spaces of object part geometries, which we can use to fit partial, real-world RGB-D scans of a scene, decomposing detected objects into their complete part geometries.
             In contrast to 3D scene understanding approaches that make independent predictions per-object, our parametric part spaces enables formulating test-time constraints for consistency within an input scene, thus producing both accurate as well as globally-consistent part decompositions. 
     		%Initially, objects are detected as bounding boxes, and for each detected object we estimate its semantic part labels and map them into the latent part space.
             %Focusing on test-time, we can then optimize the part representations to fit to the input scan; furthermore, rather than independently optimizing each object, we optimize for consistency with similar objects in the same scene, producing globally-consistent, complete part decompositions.
 		}
             \vspace{-0.1cm}
 		\label{fig:teaser}
 	\end{center}
 }]

\maketitle

\begin{abstract}
3D scene understanding has seen significant advances in recent years, but has largely focused on object understanding in 3D scenes with independent per-object predictions.
%mostly focusing on making predictions as single forward passes, while neglecting to learn the efficiently structured and rich latent spaces. 
We thus propose to learn \OURSFULL{} (\OURS{}), parametric spaces of objects and their parts, that enable optimizing to fit to a new input 3D scan with global scene consistency constraints. 
The rich structure of our \OURS{} enables accurate, holistic scene reconstruction across similar objects in the scene. 
Both objects and their part geometries are characterized by coordinate field MLPs, facilitating optimization at test time to fit to input geometric observations as well as similar objects in the input scan. 
This enables more accurate reconstructions than independent per-object predictions as a single forward pass, while establishing global consistency within a scene.
Experiments on the ScanNet dataset demonstrate that \OURS{} significantly outperforms the state-of-the-art in part decomposition and object completion in real-world scenes.

{\normalfont Project page:} {\normalfont \url{ alexeybokhovkin.github.io/neural-part-priors/}}
\end{abstract}

\section{Introduction}
\label{sec:intro}

With the introduction of commodity RGB-D sensors (e.g., Microsoft Kinect, Intel RealSense, etc.), remarkable progress has been made in reconstruction and tracking to construct 3D models of real-world environments \cite{newcombe2011kinectfusion,whelan2015elasticfusion,choi2015robust,niessner2013real,dai2017bundlefusion}.
This has enabled construction of large-scale datasets of real-world 3D scanned environments \cite{dai2017scannet,Matterport3D}, enabling significant advances in 3D semantic segmentation \cite{dai20183dmv,graham20183dsemantic,choy20194d}, 3D semantic instance segmentation \cite{hou20193dsis,engelmann20203d,han2020occuseg}, and even part-level understanding of scenes~\cite{bokhovkin2021towards}.
The achieved 3D object recognition has shown impressive advances, but methods focus on independent predictions per object in single forward passes, resulting in semantic predictions that are inconsistent between repeated objects in a scene, and/or geometric predictions that do not precisely match input observed geometry.

Simultaneously, recent advances in representing 3D shapes as continuous implicit functions represented with coordinate field MLPs have shown high-fidelity shape reconstruction \cite{park2019deepsdf, chibane2020ifnet, chen2019imnet, sitzmann2019siren, chibane2020ndf}.
Such methods have focused on object-level reconstructions, whereas part-based understanding is fundamental to many higher-level scene understanding tasks (e.g., interactions often occur with object parts -- sitting on the seat of a couch, opening a door with a handle, etc.).

%Simultaneously, significant work has been done towards part segmentation on 3D shapes, where synthetic datasets are available with part annotations for supervision \cite{mo2019partnet}. Recently, Bokhovkin et al.~\cite{bokhovkin2021towards} proposed to bridge two tasks together to predict part decompositions of objects in real-world 3D scenes, leveraging part information from synthetic CAD models aligned to real-world scanned scenes \cite{avetisyan2019scan2cad,shapenet2015,dai2017scannet}. This made an important step towards part-based understanding in 3D scenes. However, due to a single forward pass prediction, the aforementioned methods and Bokhovkin et al. focus on independent per-object prediction and are not able to reconstruct every shape in a dataset with acceptable accuracy on average, but not in particular, as optimization-based methods do. These approaches are not able to keep enough accuracy between predicted geometry and the actual underlying scan geometry and a sufficient level of consistency between predicted shapes belonging to one scene. Nevertheless, the greatest drawbacks of the latter are long runtime and the requirement of proper initialization. In addition, Bokhovkin et al.~\cite{bokhovkin2021towards} treats each object on a scene independently and is not able to achieve high accuracy surface reconstruction due to a dense volumetric representation, relying heavily on synthetic priors that did not match precisely to real observations.

We thus propose to learn \OURSFULL{} (\OURS{}), optimizable parametric shape and part spaces learned from synthetic data.
These learned manifolds enable efficient traversal in latent spaces during inference to fit precisely to objects in real-world scanned scenes, while maintaining consistent part decompositions with similar objects in the scene. 
Our \OURS{} leverage the representation power of neural implicit functions encoded as coordinate-field MLPs, representing both shape and part geometries of objects.
A shape can then be represented by a set of latent codes for each of its parts, where each code decodes to predict the respective part segmentation and signed distance field representation of the part geometry.
This representation enables effective \textit{test-time joint optimization} over all parts of a shape by traversing through the part latent space to find the set of parts that best explain a real-world shape observation.
Furthermore, as repeated objects often appear in a scene under different partial observation patterns (resulting in inconsistent predictions when made independently for each object) we further optimize for part consistency between similar objects detected in a scene to produce scene-consistent part decompositions. This allows us to reconstruct the holistic structure of a scene. 

To fit real-world 3D scan data, we first perform object detection and estimate the part types for each detected object.
We can then optimize \textit{in test time} jointly over the part codes for each shape to fit the observed scan geometry; we leverage a predicted part segmentation of the detected object and optimize jointly across the parts of each shape such that each part matches the segmentation, and their union fits the object.
This joint optimization across parts produces a high-resolution part decomposition whose union represents the complete shape while fitting precisely to the observed real geometry.
Furthermore, this optimization at inference time allows leveraging global scene information to inform our optimized part decompositions; in particular, we consider objects of the same predicted class with similar estimated geometry, and optimize them jointly, enabling more robust and scene-consistent part decompositions.

%TODO HERE

In summary, we present the following contributions:
\begin{itemize}[leftmargin=*]
\setlength\itemsep{-.3em}
\item We propose to learn optimizable part-based latent priors for 3D shapes -- \OURSFULL{}, encoding part segmentation and part geometry into a latent space for the part types of each class category.

\item Our learned, optimizable part priors enable test-time optimization over the latent spaces, enhanced with \textit{inter-shape part-based constraints}, to fit partial, cluttered object geometry in real-world scanned scenes, resulting in robust and precise semantic part completion.

\item We additionally propose a scene-consistent optimization, enhanced with \textit{intra-shape constraints}, jointly optimizing over similar objects that provides globally-consistent part decompositions for repeated object instances in a scene.
\end{itemize}
\vspace{-0.3cm}
\section{Related Works}
\label{sec:related}
\vspace{-0.3cm}

\paragraph{3D Object Detection and Instance Segmentation}
3D semantic scene understanding has seen rapid progress in recent years, with large-scale 3D datasets \cite{dai2017scannet,Matterport3D,geiger2012we} and developments in 3D deep learning showing significant advances in object-level understanding of 3D scenes.
Various methods explore learning on different 3D representations for object detection and instance segmentation, including volumetric grids \cite{hou20193dsis,tung2019learning}, point clouds \cite{qi2019deep,xie2020mlcvnet,jiang2020pointgroup,qi2020imvotenet,nie2021rfd}, sparse voxel representations \cite{engelmann20203d,han2020occuseg}, and hybrid approaches \cite{hou20193dsis,qi2020imvotenet}.
These approaches have achieved impressive performance in detecting and segmenting objects in real-world observations of 3D scenes, but do not consider lower-level object part information that is requisite for many vision and robotics tasks, that involve object interaction and manipulation.

Recently, Bokhovkin et al.~\cite{bokhovkin2021towards} proposed an approach to estimate part decompositions of objects in RGB-D scans, leveraging structural part type prediction and a pre-computed set of geometric part priors.
Due to the use of dense volumetric part priors, the part reasoning is limited to coarse resolutions, and does not precisely match the input observed geometry.
We also address the task of semantic part prediction and completion for objects in real-world 3D scans, but leverage a learned, structured latent space representing neural part priors, enabling part reasoning at high resolutions which optimizing to fit accurately to the observed scan geometry.

\paragraph{3D Scan Completion}
As real-world 3D reconstructions are very often incomplete due to the complexity of the scene geometry and occlusions, various approaches have been developed to predict complete shape or scene geometry from partial observations.
Earlier works focused on voxel-based scan completion for shapes \cite{wu20153d,dai2017shape}, with more recent works tackling the challenge of generating complete geometry from partial observations of large-scale scenes \cite{song2017semantic,dai2018scancomplete,dai2020sgnn,dai2021spsg}, but without considering individual object instances.
Several recent works propose to detect objects in an RGB-D scan and estimate the complete object geometries, leveraging voxel \cite{hou2020revealnet,bokhovkin2021towards} or point \cite{yi2019gspn,nie2021rfd} representations.
Our approach to predicting part decompositions of objects inherently provides object completion as the predicted parts union; in contrast to previous approaches estimating object completion in RGB-D scans, we propose to characterize object parts as learned implicit priors, enabling test-time traversal of the latent space to fit accurately to observed input scan geometry.

\paragraph{Part Segmentation of 3D Shapes}
Part segmentation for 3D shapes has been well-studied in shape analysis, typically focusing on understanding collections of synthetic shapes.
Various methods have been developed for unsupervised part segmentation by finding a consistent segmentation across a set of shapes \cite{golovinskiy2009consistent,huang2011joint,sidi2011unsupervised,hu2012co,luo2020learning}.
Recent deep learning based approaches have leveraged datasets of shapes with part annotations to learn part segmentation on new shapes \cite{kalogerakis20173d,yi2016scalable,hanocka2019meshcnn}.
In particular, approaches that learn part sequences and hierarchies to capture part structures have shown effective part segmentation for shapes \cite{wang2011symmetry,van2013co,yi2017learning,mo2019partnet,mo2019structurenet,wu2020pq}.
These approaches target single-object scenarios, whereas we construct a set of learned part priors that can be optimized to fit to real-world, noisy, incomplete scan geometry.

\paragraph{Neural Implicit Representations of 3D Shapes}
Recently, we have seen significant advances in generative shape modeling with learned neural implicit representations that can represent continuous implicit surface representations, without ties to an explicit grid structure.
Notably, DeepSDF~\cite{park2019deepsdf} proposed an MLP-based network that predicts the SDF value for a given 3D location in space, conditioned on a latent shape code, which demonstrated effective modeling of 3D shapes while traversing the learned shape space.
Such implicit representations have also been leveraged in hybrid approaches coupling explicit geometric locations with local implicit descriptions of geometry for shapes \cite{genova2019learning,genova2020local} as well as scenes \cite{peng2020convolutional}, without semantic meaning to the local decompositions. 
We propose to leverage the representation power of such learned continuous implicit surfaces to characterize semantic object parts that can be jointly optimized together to fit all parts of an object to a partial scan observation.

\section{Method}

% Overview
\begin{figure*}
\begin{center}
    \centering
    \includegraphics[width=0.85\textwidth]{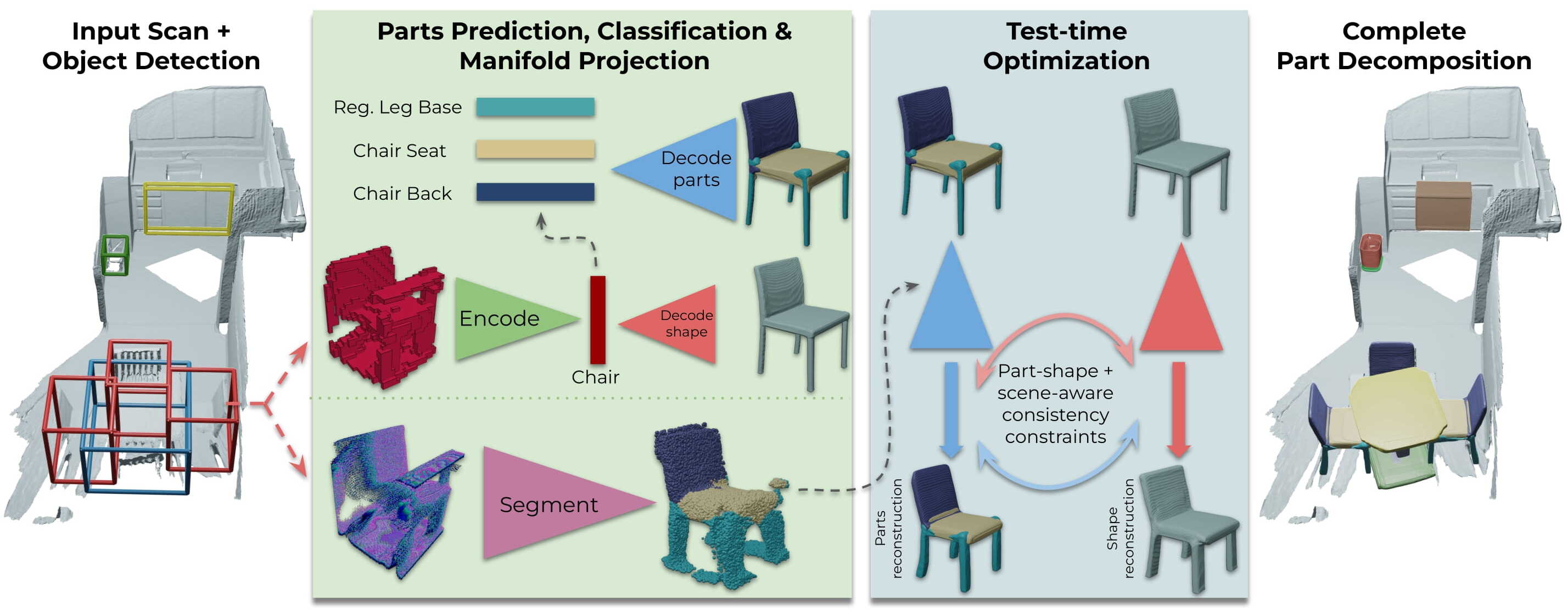}
    \vspace{-0.3cm}
    \caption{Method overview.
    From an input scan, we first detect 3D bounding boxes for objects. 
    For each object, we predict their semantic part structure as a set of part labels and latent codes for each part. 
    These latent codes map into the space of neural part priors, along with a full shape code used to regularize the shape structure.
    We  then refine these codes at test time by optimizing to fit to the observed input geometry along with inter-object consistency between similar detected objects, producing effective part decompositions reflecting complete objects with scene consistency.
    \vspace{-0.9cm}
    }
    \label{fig:overview}
\end{center}%
\end{figure*}

% % Overview
% \begin{figure}
% \begin{center}
%     \centering
%     \includegraphics[width=1.0\textwidth]{fig/overview_1.jpg}
%     \caption{ \TODO{caption}
%     }
%     \label{fig:overview_1}
% \end{center}%
% \end{figure}

% % Overview
% \begin{figure}
% \begin{center}
%     \centering
%     \includegraphics[width=1.0\textwidth]{fig/overview_2.jpg}
%     \caption{ \TODO{caption}
%     }
%     \label{fig:overview_2}
% \end{center}%
% \end{figure}

\subsection{Overview}

We introduce \OURSFULL{} (\OURS{}) to represent learned spaces of geometric object part priors, that enable joint part segmentation and completion of objects in real-world, incomplete RGB-D scans.
From an input 3D scan $\mathcal{S}$, we first detect objects $\mathcal{O}$ = \{$o_i$\} in the scan characterized by their bounding boxes and orientations, then for each object, we predict its part decomposition into a part class categories (with corresponding part latent codes) and their corresponding complete geometry represented as signed distance fields (SDFs) and trained on part annotations for shapes.
This enables holistic reasoning about each object in the scene and prediction of complete geometry in unobserved scan regions.
Since captured real-world scene geometry contains significant incompleteness or noise, we  model our geometric part priors based on complete, clean synthetic object parts, represented as a learned latent space over implicit part geometry functions.
This enables test-time optimization over the latent space of parts to fit real geometry observations, enabling part-based object completion while precisely representing real object geometry.
Rather than considering each object independently, we observe that repeated objects often occur in scenes under different partial observations, leading to inconsistent independent predictions; we thus jointly optimize across similar objects in a scene, where objects are considered similar if they share the same class category and predicted shape geometries are close by chamfer distance. This results in scene-consistent, high-fidelity characterizations of both object part semantics and complete object geometry.
%\ALEX{During test-time optimization (TTO) we also incorporate two types of consistency constraints. First, we force optimization of parts geometry in TTO to be guided with an optimization of full shape geometry. This helps to keep completeness of every optimized part and consistency between parts. Second, we make use of similar instances on a scene with our scene-aware consistency constraints. We identify similar objects on a scan that have the same shape category and aggregate information from all of them to produce a unified shape.}
An overview of our approach is shown in Fig.~\ref{fig:overview}. 

% Our \OURS{} spaces characterize object part geometry as signed distance fields (SDFs), trained on part annotations for shapes.
% For each detected object $o_i\in \mathcal{O}$ in an input scan, we predict its semantic parts as the set of part categories that compose the object along with initial estimates of their corresponding latent codes in the learned part space.
% We then optimize for refined object pose alignment, and then for the part latent codes to fit to the observed geometry of each $o_i$.
% For objects of the same class category and with similar predicted shape geometry by chamfer distance, we jointly constrain them together in this optimization.
% This results in scene-consistent, high-fidelity characterizations of both object part semantics and complete object geometry.

\subsection{Object Detection}
From input 3D scan $\mathcal{S}$, we first detect objects in the scene, leveraging a state-of-the-art 3D object detection backbone from MLCVNet~\cite{xie2020mlcvnet}.
MLCVNet interprets $\mathcal{S}$ as a point cloud and proposes objects through voting~\cite{qi2019deep} at multiple resolutions, providing an output set of axis-aligned bounding boxes for each detected object $o_i$.
We extract the truncated signed distance field $D_i$ for each $o_i$ at $4$mm resolution to use for test-time optimization.
We then aim to characterize shape properties for $o_i$ to be used for rotation estimation and test-time optimization, and interpret $D_i$ as a $32^3$ occupancy grid which is input to a 3D convolutional object encoder to produce the object's shape descriptor $\mathbf{s}_i\in\mathbb{R}^{256}$.
%\ANGIE{we extract both 32 occupancy and high-res sdf? the occupancy for the model and the sdf for test time fitting instead of sdf as input to the model?} \ALEX{Yes, we extract first high-res sdf with resolution of 4 mm, then convert this field to occupancy by threshold $abs(sdf) < 0.02$}

\paragraph{Initial Rotation Estimation} From $\mathbf{s}_i$, we use a 2-layer MLP to additionally predict an initial rotation estimate of the object as $r_i^{\mathrm{init}}$ around the up (gravity) vector of $\mathcal{S}$.
We note that the up vectors of an RGB-D scan can be reliably estimated with IMU and/or floor estimation techniques \cite{dai2017scannet}.
The rotation estimation is treated as a classification problem across $n_{r}=12$ bins of discretized angles ($\{0^{\circ}, 30^{\circ}, \dots, 330^{\circ}\}$), using a cross entropy loss.
We use the estimated rotation $r_i^{\mathrm{init}}$ to resample $D_i$ to approximate the canonical object orientation, from which we use to optimize for the final rotation $r_i$ and the object part latent codes.

%We then resample the input scan geometry within each detected box into $32^3$ occupancy grids $o_i\in \mathbb{O}$ to inform our part decomposition. In addition, we extract a truncated signed distance field (tSDF) $sdf_i$ within predicted bounding box, where truncation distance is $0000$ and tSDF resolution is $0000$ cm. Truncated SDFs are obtained with fusion of depthmaps.
%For a detected object $o_i$ from the scan, represented as a $32^3$ occupancy grid of the scan geometry within its predicted bounding box, we encode the occupancy grid with  four 3D convolutional blocks (consisting of convolution, group normalization and ReLU activation) and extract a feature encoding $z_i$ of dimension $256$, which is used to inform the part decomposition. 

%\paragraph{Object Orientation Prediction}
%Since our object detection backbone predicts axis-aligned bounding boxes for each object, we additionally predict the orientation $r_i$ of each object $o_i$ from its feature $z_i$ using an MLP.
%We assume that the up (gravity) vector is known in the scene, and thus predict the angle around the up vector by classifying the angle in $n_{\alpha}=12$ bins of discretized angles ($\{0^{\circ}, 30^{\circ}, \dots, 330^{\circ}\}$) with a cross entropy loss.
%The predicted object orientation allows us to initialize the orientation of input signed distance field $sdf_i$ for the following ICP registration, decribed in Section~\ref{}.

\subsection{Learned Space of \OURSFULL{}}
We first learn a set of latent part spaces for each class category, where each part space represents all part types for the particular object category.
To this end, we employ a function $f_p$ characterized as an MLP to predict the implicit signed distance representation for each part geometry of the class category.
In addition to the latent part space, we additionally train a proxy shape function $f_s$ as an MLP that learns full shape geometry as implicit signed distances, which will serve as additional regularization during the part optimization.
Both $f_p$ and $f_s$ are trained in auto-decoder fashion following DeepSDF~\cite{park2019deepsdf}.
Then each train shape part is embedded into a part latent space by optimizing for its code  $\mathbf{z}^{\mathrm{p}}_k\in \mathbb{R}^{256}$ such that $f_p$ conditioned on this code and the part type maps a positional encoding $\mathbf{x}^{pos}\in\mathbb{R}^{63}$ of a point $\mathbf{x}\in\mathbb{R}^3$ in the canonical space to SDF value $d$ of the part geometry:
\begin{equation}
    f_p: \mathbb{R}^{63} \times \mathbb{R}^{256}\times \mathbb{Z}_2^{N_c} \rightarrow\mathbb{R}, \quad\quad f_p(\mathbf{x}^{pos},\mathbf{z}^{\mathrm{p}}_k,\mathds{1}_{\mathrm{part}})=d.
\end{equation} 
where $\mathds{1}_{\mathrm{part}}\in\mathbb{Z}_2^{N_c}$ is a one-hot encoding of the part type for a maximum of $N_c$ parts. Similar to NeRF~\cite{mildenhall2020nerf}, euclidean coordinates $\mathbf{x}\in\mathbb{R}^3$ are encoded using $\sin/\cos$ functions with 10 frequencies $[2^0,...,2^9]$.
The shape space is trained analogously for each class category where $\mathbf{z}^{\mathrm{s}}_i\in \mathbb{R}^{256}$ represents a shape latent code in the space:
%Similarly, each shape is encoded into a latent code $\mathbf{z}^{\mathrm{s}}_i\in \mathbb{R}^{256}$ \ALEX{Similarly, each shape is embedded into a DeepSDF shape latent space as a code $\mathbf{z}^{\mathrm{s}}_i\in \mathbb{R}^{256}$}, with the $f_s$ mapping a point $\mathbf{x}\in\mathbb{R}^3$ in the canonical space to an SDF value $d$, conditioned on $z^{\mathrm{s}}_i$:
\begin{equation}
    f_s: \mathbb{R}^{63} \times \mathbb{R}^{256}\rightarrow\mathbb{R}, \quad\quad f_p(\mathbf{x}^{pos},\mathbf{z}^{\mathrm{s}}_i)=d.
\end{equation}
%We try our method for several furniture categories, such as chairs, tables, beds etc. To perform efficient inference that includes optimization in test-time in implicit domain we need powerful and implicit priors for shape and part category. The DeepSDF approach is able to provide such priors, where they are represented as combinations of learned DeepSDF latent space and MLP (multi-layer perceptron) decoder. Thus, for every shape category we train $2$ types of decoders. One is trained for the whole shapes: $$s = MLP_{shape}([x, lat]),\ x \in \mathbb{R}^3,\ lat\in \mathbb{R}^{256},\ s \in \mathbb{R},$$ another one is trained for all parts of all shapes: $$s = MLP_{part}([x, lat, \mathds{1}_{part}]),\ x \in \mathbb{R}^3,\ lat\in \mathbb{R}^{256},\ s \in \mathbb{R},$$ where $x$ is a 3D point, $s$ is a corresponding distance to a surface, $lat$ is a shape or part latent vector, $\mathds{1}_{part} \in \mathbb{R}^{n^{part}_c}$ is a one-hot vector of a part type ($n^{part}_c$ -- number of parts for an arbitrary shape category $c$). 
%To pretrain these MLPs as implicit functions we follow the original DeepSDF approach, where we use ShapeNet~\cite{} dataset of synthetic shapes and split every shape on parts using PartNet~\cite{} labels. 

We train latent spaces of part and shape priors on the synthetic PartNet~\cite{mo2019partnet} dataset to characterize a space of complete parts and shapes, by minimizing the reconstruction error over all train shape parts, while optimizing for latent codes $\{ \mathbf{z}^{\mathrm{p}}_k \}$ and weights of $f_p$.
We use an $\ell_1$ reconstruction loss with $\ell_2$ regularization on the latent codes:
\begin{equation}
    L = \sum_{j=1}^{N_p} |f_p(\mathbf{x}_j^{pos},\mathbf{z}^{\mathrm{p}}_k,\mathds{1}_{\mathrm{part}}) - D^\textrm{gt}(\mathbf{x}_j^{pos})|_1 + \lambda||\mathbf{z}^{\mathrm{p}}_k||_2^2
\end{equation}
for $N_p$ points near the surface, the regularization weight $\lambda=$1e-5.
We train $f_s$ analogously.

\begin{figure*}
\centering
\begin{subfigure}{.52\textwidth}
  \centering
    \includegraphics[width=1.0\textwidth]{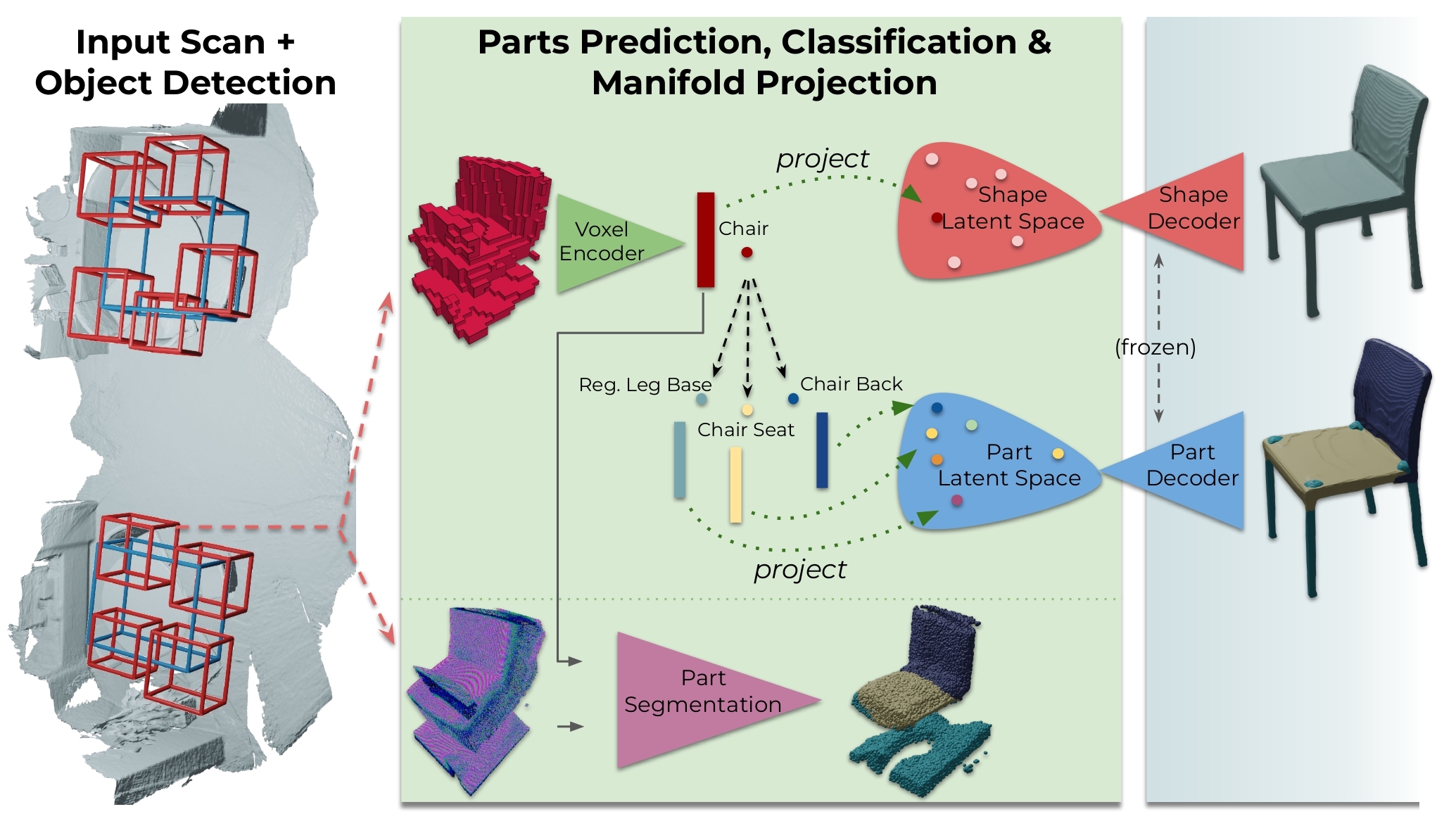}
    \caption{ Projection into part and shape spaces.
    }
    \label{fig:overview_1}
\end{subfigure}%
\begin{subfigure}{.48\textwidth}
  \centering
    \includegraphics[width=1.0\textwidth]{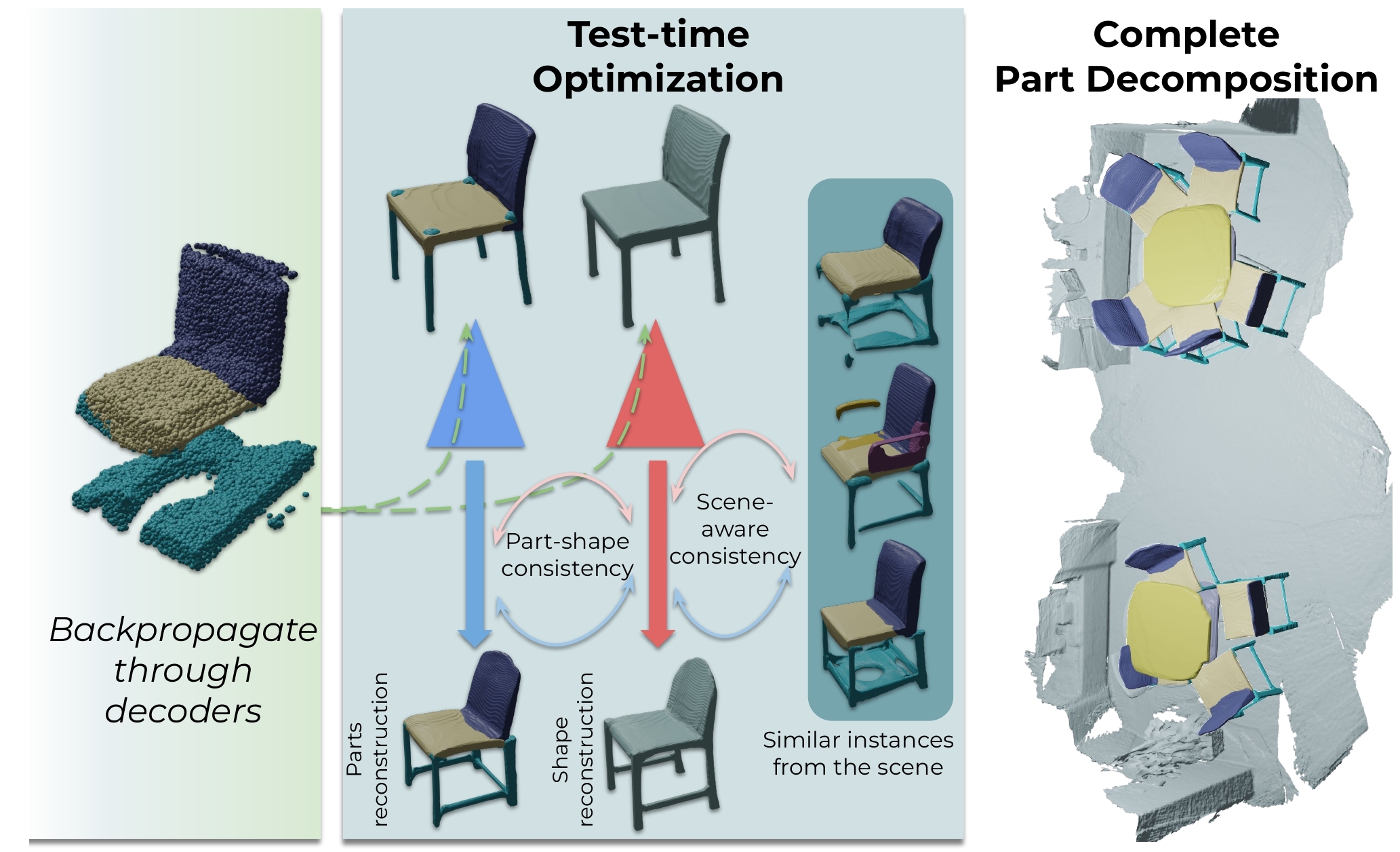}
    \caption{ Joint optimization to fit to input scan.
    }
    \label{fig:overview_2}
\end{subfigure}
\vspace{-0.2cm}
\caption{(a) Projection into the part and shape latent spaces along with part segmentation from input scan geometry. (b) Optimization at test time to fit to observed scan geometry while maintaining inter-part consistency within a shape and inter-shape consistency for geometrically similar objects.}
\vspace{-0.2cm}
\label{fig:test}
\end{figure*}

\subsection{Part Decompositions in Real Scenes}

Once we have learned our latent space of parts, we can traverse them at test time to find the part-based decomposition of an object that best fits to its real-world observed geometry in a scene.
Since real-world observations are typically incomplete, we can optimize for complete part decompositions based on strong priors given by the trained latent spaces.
This allows for effective regularization by synthetic part characteristics (clean, complete) while fitting precisely to real observed geometry.

To guide this optimization for a detected object box $o$ with its shape feature $\mathbf{s}$, we predict its high-level decomposition into a set of semantic part types $\{(c_k, \mathbf{p}_k)\}$, where $\mathbf{p}_k\in\mathbb{R}^{256}$ is a part feature descriptor and $c_k$ the part class label.
The $i$ object suffix is discarded here for simplicity.
The part optimization is initialized using $\{(c_k, \mathbf{p}_k)\}$.

To obtain the semantic part type predictions, we employ a message-passing graph neural network that infers part relations to predict the set of component part types.
Similar to \cite{mo2019structurenet}, from the shape feature $\mathbf{s}$ we use an MLP to predict at most $N_c=10$ parts.
For each potential part $k$, we predict its probability of existence, its part label $c_k$, its feature vector $\mathbf{p}_k$, and probabilities of physical adjacency (given by face connectivity between parts) between each pair of parts to learn structural part information.
This produces the semantic description of the set of parts for the object $\{(c_k, \mathbf{p}_k)\}$, from the parts predicted with part existence probability $>0.5$.

\paragraph{Projection to the Latent Part Space}
We then learn a projection mapping from the part features $\{\mathbf{p}_k\}$ to the learned latent part space based on synthetic part priors, using a small MLP to predict $\{\tilde{\mathbf{z}}_k^\textrm{p}\}$, as shown in Fig.~\ref{fig:test}(a).
This helps to provide a close initial estimate in the latent part space in order to initialize optimization over these part codes to fit precisely to the observed object geometry.
Similarly, we project the shape code $\mathbf{s}$ to the learned latent shape space with a small shape projection MLP to predict $\{\tilde{\mathbf{z}}^\textrm{s}\}$, which we use to help regularize the part code optimization.
Both of these projection MLPs are trained using MSE losses against the optimized train codes of the latent spaces.
%Both of these projection MLPs consist of 4 $512$ neuron layers with ReLU activations, and are trained using MSE losses against the optimized train codes of the latent spaces.

\paragraph{Part Segmentation Estimation}\label{subsec:pointcls}
In addition to our projection initialization, we estimate part segmentation $\{D^{\textrm{p}}\}_{\textrm{p}=1}^{N_{parts}}$ for the input object TSDF $D$ over the full volume, representing part SDF geometry in the regions predicted as corresponding to the part p, where part segmentation regions cover the entire shape, including unobserved regions.
This is used to guide part geometry predictions when optimizing at test time to fit to real observed input geometry. 
For each point $\mathbf{x}\in\mathbb{R}^3$ which has distance $<d_{trunc} = 0.16$m from the input object TSDF $D$, we classify it to one of the predicted parts $\{(c_k, \mathbf{p}_k)\}$ or background using a small PointNet-based\cite{qi2017pointnet} network.
This segmentation prediction takes as input the corresponding shape feature $\mathbf{s}$, the initial estimated rotation $r_i^{\mathrm{init}}$, and the 3D coordinates of $\mathbf{x}$, and is trained with a cross-entropy loss.

\subsection{Joint Inter- and Intra-Shape Part Optimization}\label{subsec:partopt}
To obtain the final part decompositions, we traverse over the learned latent part space to fit to the observed input scan geometry, as shown in Fig.~\ref{fig:test}(b).
From the initial estimated part codes $\{\tilde{\mathbf{z}}_k^\textrm{p}\}$ and shape code $\{\tilde{\mathbf{z}}^\textrm{s}\}$, their decoded part SDFs should match to each of $\{D^{\textrm{p}}\}_{\textrm{p}=1}^{N_{parts}}$.
Since the part and shape latent spaces have been trained in the canonical shape space, we optimize for a refined rotation prediction $r$ from $r^\textrm{init}$ using iterative closest points \cite{besl1992method,rusinkiewicz2001efficient} between the sampled points near $D$ and the initial shape estimate from projection $\tilde{\mathbf{z}}^\textrm{s}$.
We use $N_i$ sampled points near the observed input surface $D$ (near being SDF values $<0.025$m) for rotation refinement, with $N$ the number of points not predicted as background during part segmentation.
%\ALEX{We also denote $N$ as number of points not being predicted as background from $D$.}

%We then predict the final part decomposition in two stages. First, for each element $\{z'_k\}$ in the predicted semantic part arrangement $T_i$, we perform inference with DeepSDF decoders to obtain initial complete signed distance fields $\overline{sdf}^{shape}_i$ for the whole shape and $\{\overline{sdf}^{part}_1,\dots,\overline{sdf}^{part}_{n^{part}}\}$ for corresponding parts. After that we can subsample predicted fields to near-surface points $sdf'^{shape}_i$ and $\{\overline{sdf}'^{part}_1,\dots,\overline{sdf}'^{part}_{n^{part}}\}$ by taking only the points with distance less than threshold $\theta_{surf}=0.005$. The same operation is performed for the input field $sdf_i$ of a scan to get a $sdf'_i$.
%To enable efficient and accurate overfitting to a particular example in test time we perform preliminary ICP registration of the predicted $\overline{sdf}'^{shape}_i$ with the $sdf'_i$ from a scan to get a registration transformation $M$. After that we classify the points of partial field $sdf_i$ to parts $\{sdf^{part}_1,\dots,sdf^{part}_{n^{part}}\}$ using an approach of Section~\ref{subsec:pointcls} and apply transform $M$ to get $sdf'_i$ and $\{sdf'^{part}_1,\dots,sdf'^{part}_{n^{part}}\}$.

%\paragraph{Part-based Implicit Test-Time Optimization}
While the predicted projected part and shape codes $\{\tilde{\mathbf{z}}_k^\textrm{p}\},\{\tilde{\mathbf{z}}_k^\textrm{s}\}$ can provide a good initial estimate of the part decomposition of the complete shape, they represent synthetic part and shape priors that often do not fit the observed real input geometry.
We thus optimize for part decompositions that best fit the input observations by minimizing the energy:
\begin{equation}\label{eq:optimization}
    L = \sum_k L_{\textrm{part}} + L_{\textrm{shape}} + w_\textrm{cons} L_{\textrm{cons}},
\end{equation}
where $L_{\textrm{part}}$ denotes the part reconstruction loss, $L_{\textrm{shape}}$ a proxy shape reconstruction loss, $L_{\textrm{cons}}$ a regularization to encourage global part consistency within the estimated shape, and $w_\textrm{cons}$ is a consistency weight.

$L_{\textrm{part}}$ is an $\ell_1$ loss on part reconstruction:
\begin{equation}\label{eq:parts}
    L_{\textrm{part}} = \sum_{\textrm{p}=1}^{N_{parts}}\sum_{N^{\textrm{p}}} w_{trunc}|f_p(\mathbf{z}_k^\textrm{p}) - T_r(D^\textrm{p})| + \lambda||\mathbf{z}^{\mathrm{p}}_k||_2^2,
\end{equation}
where $N^{\textrm{p}}$ is the number of points classified to part $\textrm{p}$ and  $w_{trunc}$ gives a fixed greater weight for near-surface points ($<d_{trunc} = 0.16$m). % than those further away.
%\ALEX{and $w_{trunc}$ switches between points near the surface (less than $d_{trunc} = 0.16$m) than those further away}. 

$L_{\textrm{shape}}$ is a proxy $\ell_1$ loss on shape reconstruction:
\begin{equation}\label{eq:shape}
    L_{\textrm{shape}} = \sum_{N} w_{trunc}|f_s(\mathbf{z}^\textrm{s}) - T_r(D)| + \lambda||\mathbf{z}^{\textrm{s}}||_2^2.
\end{equation}

The regularization weight $\lambda=$1e-5 for both Eq.~\ref{eq:parts},~\ref{eq:shape}. Finally, $L_{\textrm{cons}}$ encourages all parts to reconstruct a shape similar to the optimized shape:
\begin{equation}\label{eq:cons}
    L_{\textrm{cons}} = \sum_{N} |f_p(\mathbf{z}_k^\textrm{p}) - f_s(\mathbf{z}^\textrm{s})|,
\end{equation}
where $f_s(\mathbf{z}^\textrm{s})$ is frozen for $L_{\textrm{cons}}$.
This allows for reconstructed parts to join together smoothly without boundary artifacts to holistically reconstruct a shape.

This produces a final optimized set of parts for each object in the scene, where parts both fit precisely to input geometry and represent the complete geometry of each part, even in unobserved regions.
The final part geometries can be extracted from the SDFs with Marching Cubes~\cite{lorensen1987marching} to obtain a surface mesh representation of the semantic part completion.

\paragraph{Scene-Consistent Optimization}
% While this formulation enables joint part optimization within an object, we observe that multiple objects within a scene often tend to correlate with each other.
Scenes often contain repeated instances of objects which are observed from different views, thus frequently resulting in inconsistent part decompositions when considered as independent objects.
We propose a scene-consistent optimization between similar predicted objects, where objects in a scene are considered similar if their predicted class category is the same and the chamfer distance between their decoded shapes from $\tilde{\mathbf{z}}_k^\textrm{s}$ is $<\tau_s$.

For a set of $N_{sim}$ similar objects in a scene, we collect together their predicted part segmentations and observed input SDF geometry in the canonical orientation based on $T_r(D^\textrm{p})$ to provide a holistic set of constraints across different partial observations to produce $\{D_i\}_{i = 1}^{N_{sim}}$. 
The $\{D_i\}_{i = 1}^{N_{sim}}$ are then aggregated to form $\mathbf{D'}$ by sampling a set of $N_{avg}$ points near the surfaces of $\{D_i\}_{i = 1}^{N_{sim}}$ where $N_{avg}$ is the average number of points across the $N_{sim}$ objects, and each point is assigned the minimum SDF value within its 30-point local neighbourhood (to help ensure that objects do not grow thicker in size from small-scale misalignments).
Based on  $\mathbf{D'}$, we then optimize for the part decompositions following Eq.~\ref{eq:optimization}.

\subsection{Implementation Details}
We first train our latent part and shape spaces on per-category on the synthetic PartNet~\cite{mo2019partnet} dataset.
We then train the projection mapping into the learned part and shape spaces as well as the part segmentation.
This is first pre-trained on synthetic PartNet data using virtually scanned incomplete inputs to take advantage of the large amount of synthetic data.
To apply to real-world observations, we then fine-tune the projections and part segmentation on ScanNet~\cite{dai2017scannet} data using MLCVNet~\cite{xie2020mlcvnet} detections on train scenes. 

In test time, we optimize for part and shape codes using an Adam optimizer with learning rate of 3e-4 for 300 iterations and 3e-5 for the next 300 iterations.
To enable more flexibility to capture input details, we optimize the decoder weights for parts and shape after 400 iterations. 
Optimization for each part takes $\approx 25$ seconds.
For further implementation and training details, we refer to the supplemental.
\begin{table*}[tp]
\centering
\resizebox{\textwidth}{!}{
\vspace{-0.5cm}
\begin{tabular}{l|cccccc|cc||cccccc|cc}
& \multicolumn{8}{c||}{Chamfer Distance -- Accuracy ($\downarrow$)} & \multicolumn{8}{c}{Chamfer Distance -- Completion ($\downarrow$)}\\
\toprule
    Method & chair & table & cab. & bkshlf & bed & bin & class avg & inst avg & chair & table & cab. & bkshlf & bed & bin & class avg & inst avg \\
\midrule
    SG-NN\cite{dai2020sgnn} + MLCVNet\cite{xie2020mlcvnet} + PointGroup\cite{jiang2020pointgroup} & 0.047 & 0.110 & 0.146 & 0.173 & 0.350 & 0.051 & 0.146 & 0.083 & 0.054 & 0.141 & 0.123 & 0.192 & 0.382 & 0.045 & 0.156 & 0.089 \\
    MLCVNet\cite{xie2020mlcvnet} + StructureNet\cite{mo2019structurenet} & 0.024 & 0.074 & 0.104 & 0.166 & 0.424 & 0.039 & 0.138 & 0.061 & 0.028 & 0.129 & 0.118 & 0.154 & 0.352 & 0.037 & 0.136 & 0.067 \\
    Bokhovkin et al.\cite{bokhovkin2021towards} & 0.029 & 0.073 & 0.099 & 0.168 & 0.244 & 0.036 & 0.108 & 0.056 & 0.031 & 0.095 & \textbf{0.108} & \textbf{0.151} & 0.236 & 0.038 & 0.110 & 0.059 \\
\midrule
    {\bf Ours} & \textbf{0.013} & \textbf{0.049} & \textbf{0.080} & \textbf{0.134} & \textbf{0.139} & \textbf{0.022} & \textbf{0.074} & \textbf{0.035} & \textbf{0.020} & \textbf{0.073} & 0.110 & 0.193 & \textbf{0.132} & \textbf{0.023} & \textbf{0.092} & \textbf{0.045} \\
\bottomrule
\end{tabular}
}
\vspace{-0.3cm}
\caption{Evaluation of semantic part completion on Scan2CAD~\cite{avetisyan2019scan2cad} in comparison to state-of-the-art part segmentation \cite{jiang2020pointgroup,mo2019structurenet} and semantic part completion \cite{bokhovkin2021towards}.
Our optimizable part priors produce more accurate part decompositions.
}
\vspace{-0.4cm}
\label{tab:part_cmpl_comparison}
\end{table*}

\begin{table*}[bp]
\centering
\resizebox{\textwidth}{!}{
\vspace{-0.5cm}
\begin{tabular}{l|cccccc|cc||cccccc|cc}
& \multicolumn{8}{c||}{Chamfer Distance ($\downarrow$)} & \multicolumn{8}{c}{IoU ($\uparrow$)}\\
\toprule
    Method & chair & table & cab. & bkshlf & bed & bin & class avg & inst avg & chair & table & cab. & bkshlf & bed & bin & class avg & inst avg \\
\midrule
    SG-NN\cite{dai2020sgnn} + MLCVNet\cite{xie2020mlcvnet} + PointGroup\cite{jiang2020pointgroup} & 0.056 & 0.121 & 0.110 & 0.161 & 0.406 & 0.034 & 0.148 & 0.088 & 0.257 & 0.390 & \textbf{0.300} & 0.229 & 0.306 & \textbf{0.488} & 0.328 & 0.293 \\
    MLCVNet\cite{xie2020mlcvnet} + StructureNet\cite{mo2019structurenet} & 0.025 & 0.101 & 0.092 & \textbf{0.090} & 0.359 & 0.041 & 0.118 & 0.059 & 0.480 & 0.356 & 0.195 & 0.331 & 0.267 & 0.342 & 0.329 & 0.413 \\
    Bokhovkin et al.\cite{bokhovkin2021towards} & 0.027 & 0.059 & 0.095 & 0.102 & 0.207 & 0.031 & 0.087 & 0.049 & \textbf{0.548} & \textbf{0.542} & 0.224 & 0.328 & 0.442 & 0.375 & 0.409 & \textbf{0.490} \\
\midrule
    {\bf Ours} & \textbf{0.021} & \textbf{0.051} & \textbf{0.076} & 0.094 & \textbf{0.141} & \textbf{0.025} & \textbf{0.068} & \textbf{0.038} & 0.489 & \textbf{0.542} & 0.272 & \textbf{0.386} & \textbf{0.476} & 0.340 & \textbf{0.416} & 0.461 \\
\bottomrule
\end{tabular}
}
\vspace{-0.3cm}
\caption{Evaluation of part segmentation on Scan2CAD~\cite{avetisyan2019scan2cad}. We evaluate part segmentation only on observed scan geometry, in comparison with state-of-the-art part segmentation \cite{jiang2020pointgroup,mo2019structurenet} and semantic part completion \cite{bokhovkin2021towards}. }
\label{tab:part_seg_comparison}
\end{table*}

\section{Results}
\label{sec:results}

% Close-ups
\begin{figure*}
\begin{center}
    \centering
    \includegraphics[width=0.85\textwidth]{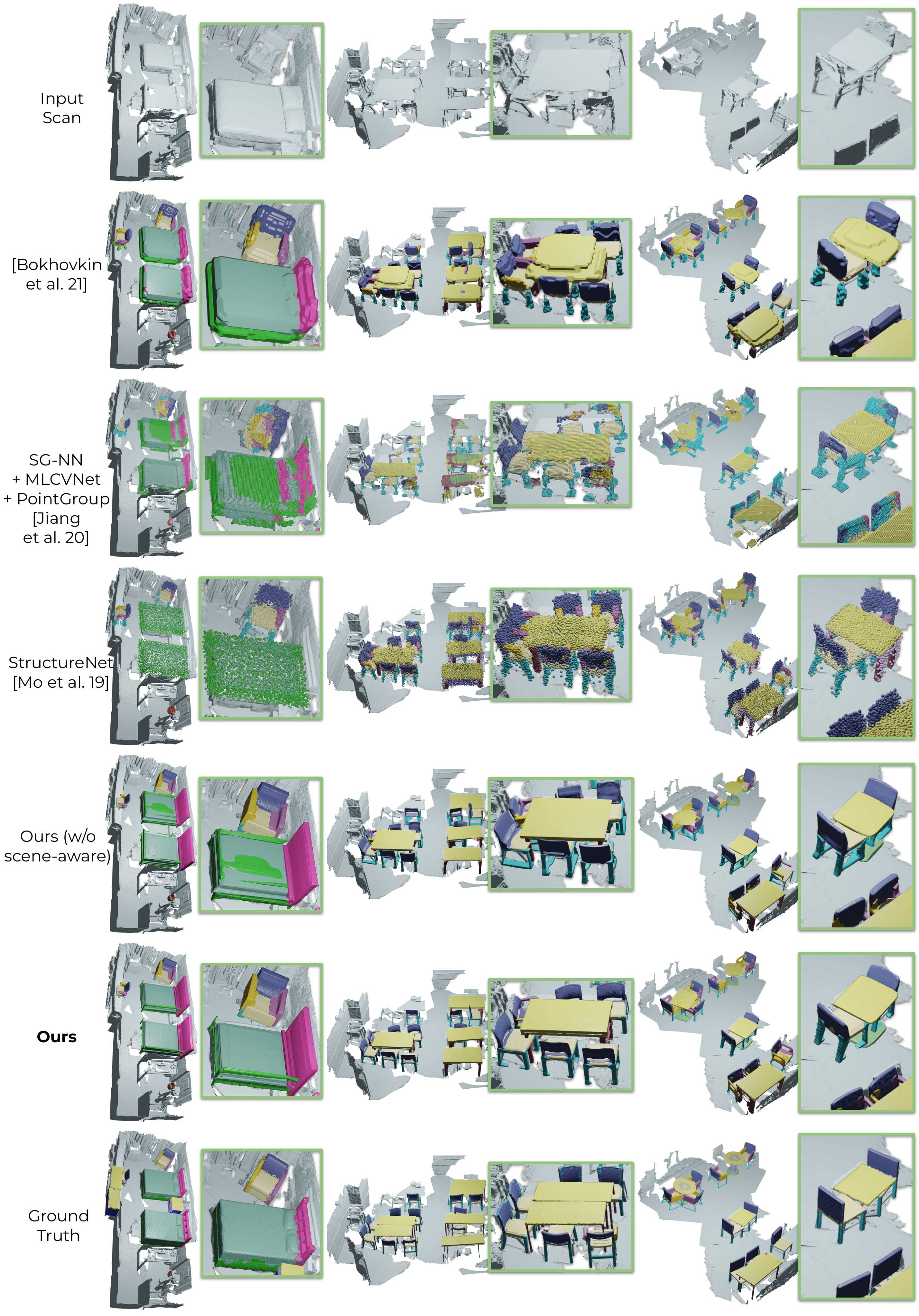}
    \vspace{-0.2cm}
    \caption{Qualitative comparison of \OURS{} with point \cite{jiang2020pointgroup,mo2019structurenet} and voxel-based \cite{bokhovkin2021towards} state of the art on ScanNet scans with Scan2CAD+PartNet ground truth. Our joint optimization across part priors enables more consistent, accurate part decompositions.}
    \label{fig:comparisons}
\end{center}%
\end{figure*}

We evaluate our \OURSFULL{} for semantic part completion on real-world RGB-D scans from ScanNet~\cite{dai2017scannet}.
We use the official train/val/test split of 1045/156/ 312 scans.
In order to evaluate part segmentation in these real-world scenes, we use the Scan2CAD~\cite{avetisyan2019scan2cad} annotations of CAD model alignments from ShapeNet~\cite{shapenet2015} to these scenes, and the PartNet~\cite{mo2019partnet} part annotations for the ShapeNet objects.
To construct our part latent space, we train on PartNet and train a projection from train ScanNet objects to their PartNet annotations; we also train all baselines on the same ScanNet+PartNet data.
We consider 6 major object class categories representing the majority of parts, comprising a total of 28 part types.
All methods are provided the same MLCVNet bounding boxes, which contain at least $40\%$ of the closest annotated shape from Scan2CAD\cite{avetisyan2019scan2cad} dataset.

\paragraph{Evaluation Metrics}
Since PartNet shapes aligned to real-world ScanNet geometry do not have precise geometric alignments, due to inexact synthetic-real associations, our evaluation metrics aim to characterize the quality of part decompositions that fit well to the real-world geometry, as well as accurately represent complete object geometry.
We evaluate \emph{accuracy} of the part segmentation with respect to the ScanNet object, and object \emph{completion} with respect to the complete object geometry from PartNet.

\emph{Accuracy} measures part segmentation predictions with respect to PartNet labels projected onto ScanNet object geometry, as a single-sided chamfer distance from the partial ScanNet object to the predicted part decomposition (as we lack complete real-world geometry available for evaluation in the other direction).
\emph{Completion} evaluates a bi-directional Chamfer distance between the predicted part decomposition against the PartNet labeled shape.

For evaluation, we sample $10,000$ points per part from predicted and ground-truth mesh surfaces and transform them to the ScanNet coordinate space.
% We consider all evaluation metrics across each part category corresponding to an object class, and evaluate class average over the categories, and instance average over all shapes. 
Each shape instance is evaluated over a union of predicted and ground-truth parts and then summed to represent shape evaluation. All shape evaluations are averaged to form instance average and the mean of shape category averages results in class average.
% If any predicted or ground-truth part does not correspond to the other, we use the center of the respective shape for evaluation.

To evaluate part segmentation of only the observed input geometry without considering completion, we use Chamfer distance and IoU. 
We project part labels from aligned PartNet shapes and predicted mesh parts onto the ScanNet mesh surface to obtain ground truth and predicted part segmentations. 
% Predicted mesh part labels are projected to ScanNet to obtain predicted part segmentation.
% We similarly use $10,000$ sampled points per part for these metrics.
For further details, we refer to the supplemental.

\vspace{-0.3cm}
\begin{table*}[bp]
\centering
\resizebox{\textwidth}{!}{
\begin{tabular}{l|cccccc|cc||cccccc|cc}
& \multicolumn{8}{c||}{Chamfer Distance ($\downarrow$) -- Accuracy} & \multicolumn{8}{c}{Chamfer Distance ($\downarrow$) -- Completion}\\
\toprule
    Method & chair & table & cab. & bkshlf & bed & bin & class avg & inst avg & chair & table & cab. & bkshlf & bed & bin & class avg & inst avg \\
\midrule
    w/o Projection Map. & 0.032 & 0.156 & 0.140 & 0.249 & 0.352 & 0.029 & 0.160 & 0.081 & 0.026 & 0.144 & 0.137 & 0.192 & 0.309 & 0.025 & 0.139 & 0.070 \\
    w/o Synthetic Pretrain & 0.026 & 0.079 & 0.116 & 0.158 & 0.152 & 0.035 & 0.094 & 0.052 & 0.029 & 0.097 & 0.132 & 0.206 & 0.234 & 0.033 & 0.122 & 0.061 \\
    w/o Test-Time Opt. & 0.019 & 0.060 & 0.105 & 0.145 & 0.200 & \underline{0.023} & 0.093 & 0.047 & \underline{0.022} & 0.085 & 0.127 & \underline{0.183} & 0.198 & \underline{0.024} & 0.107 & 0.052 \\
    w/o Full-shape Constr. & \underline{0.016} & 0.051 & 0.090 & 0.151 & 0.168 & 0.025 & 0.083 & 0.040 & 0.023 & 0.075 & 0.121 & 0.202 & 0.157 & 0.026 & 0.101 & \underline{0.050} \\
    w/o Segmentation & \textbf{0.013} & \textbf{0.046} & \textbf{0.065} & \textbf{0.120} & \textbf{0.062} & \underline{0.023} & \textbf{0.058} & \textbf{0.030} & 0.035 & 0.084 & 0.118 & 0.195 & 0.203 & 0.034 & 0.112 & 0.061 \\
    w/o Scene Consistency & \underline{0.016} & 0.053 & \underline{0.080} & 0.139 & 0.152 & \underline{0.023} & 0.077 & 0.038 & \textbf{0.020} & \underline{0.074} & \textbf{0.108} & \textbf{0.180} & \underline{0.150} & 0.025 & \underline{0.093} & \textbf{0.045} \\
    {\bf Ours} & \textbf{0.013} & \underline{0.049} & \underline{0.080} & \underline{0.134} & \underline{0.139} & \textbf{0.022} & \underline{0.074} & \underline{0.035} & \textbf{0.020} & \textbf{0.073} & \underline{0.110} & 0.193 & \textbf{0.132} & \textbf{0.023} & \textbf{0.092} & \textbf{0.045} \\
\bottomrule
\end{tabular}
}
\vspace{-0.3cm}
\caption{Ablation study evaluating semantic part completion on Scan2CAD~\cite{avetisyan2019scan2cad}. 
We show the effect of our projection mapping initialization, and test-time optimization to fit to the for our design decisions.
Overall, using both results in the best performance in RGB-D scan part decompositions.
In bold -- the best result, underlined -- the second best result.}
\label{tab:ablations}
\end{table*}

\paragraph{Comparison to state of the art}
Tab.~\ref{tab:part_cmpl_comparison} shows a comparison to state of the art on semantic part completion for real-world ScanNet scans, with qualitative results shown in Fig.~\ref{fig:comparisons}.
We compare with Bokhovkin et al.~\cite{bokhovkin2021towards}, which leverages coarse $32^3$ pre-computed geometric priors for semantic part completion, the state-of-the-art part segmentation approaches StructureNet~\cite{mo2019structurenet} and PointGroup~\cite{jiang2020pointgroup}.
Each of these methods uses the same object detection results from MLCVNet~\cite{xie2020mlcvnet}; since PointGroup does not predict any geometric completion, we provide additional scan completion from SG-NN~\cite{dai2020sgnn}.
For Bokhovkin et al.~\cite{bokhovkin2021towards} we apply Marching Cubes to voxels to extract a surface with which to evaluate.
In contrast to these approaches which estimate part decompositions directly and independently, our \OURS{} enable joint optimization over parts to fit precisely to the observed scan geometry, resulting in improved accuracy and completion performance.
Additional qualitative results are shown in the supplemental.

\paragraph{Part segmentation on 3D scans}
We also evaluate part segmentation in Tab.~\ref{tab:part_seg_comparison}, which considers segmentation of only the observed scan geometry, without any geometric completion.
We intersect each method's part predictions with the original scan geometry to evaluate this part segmentation, in comparison with Bokhovkin et al.~\cite{bokhovkin2021towards}, StructureNet~\cite{mo2019structurenet}, and PointGroup~\cite{jiang2020pointgroup}.
By jointly optimizing over the parts of an object to fit to its observed scan geometry, our approach improves notably in part segmentation performance.

%\paragraph{Ablations}
\smallskip
\noindent \textbf{What is the effect of scene-consistent optimization?}
Tab.~\ref{tab:ablations} and Fig.~\ref{fig:comparisons} show the effect of scene-consistent optimization. 
This improves results over \emph{w/o Scene Consistency}, with a stronger effect on categories that more often have repeated instances (e.g., chair, table, bed in offices, classrooms, hotel rooms).
This more holistic optimization enables more consistent reasoning about objects and their parts in a scene.

% General scenes
% \begin{figure*}
% \begin{center}
%     \centering
%     \includegraphics[width=0.95\textwidth]{fig/add_results.jpg}
%     \vspace{-0.3cm}
%     \caption{Additional qualitative results on ScanNet\cite{dai2017scannet} with Scan2CAD\cite{avetisyan2019scan2cad} and PartNet\cite{mo2019partnet} targets, showing our consistent, complete part decompositions.}
%     \label{fig:scenes}
%     \vspace{-0.9cm}
% \end{center}%
% \end{figure*}

\smallskip
\noindent \textbf{What is the impact of test-time optimization and projection initialization?}
We consider our approach without using a learned projection mapping to the latent part space as initialization for test-time optimization to fit the observed scan geometry and also evaluate only the projection mapping without test-time optimization, shown in Tab.~\ref{tab:ablations}.
The initial projection helps significantly to obtain a good initialization for test-time optimization (\emph{w/o Projection Map}), while the projection results provide a good estimate but imprecise fit to the observed scene geometry (\emph{w/o Test-Time Opt}). 
% By leveraging TTO and projection initialization, we can achieve the best representation of the input scan as its part decomposition.

\smallskip
\noindent \textbf{What is the effect of segmentation and full-shape constraints?}
The effects of full-shape constraints (\emph{w/o Full-shape Constr.}) and dense segmentation (\emph{w/o Segmentation}) are evaluated in Tab.~\ref{tab:ablations}. The full-shape constraint helps to maintain the global shape consistency of the optimized parts during TTO. Dense segmentation guides the TTO optimization constraints (e.g., avoids fitting to clutter; prevents self-intersections between optimized reconstructed parts).

% that one is less important - can move to supp
% \smallskip
% \noindent \textbf{Effect of synthetic pre-training.} 
% We evaluate the effect of synthetic pre-training of the part segmentation and projection mappings to the part and shape latent spaces in Tab.~\ref{tab:ablations} (\emph{w/o Synthetic Pretrain}).
% The additional quantity and diversity of data helps to avoid overfitting to more limited real data.

\paragraph{Limitations}
While our \OURS{} shows strong promise towards accurate, high resolution characterization of semantic parts in real-world scenes required for finer-grained semantic scene understanding, various limitations remain.
Our part latent space is trained in its canonically oriented space, and while we can optimize for shape orientations with ICP, a joint optimization or an equivariant formulation can potentially resolve sensitivity to misaligned orientations.
Finally, our scene-consistent optimization for similar shapes in a scene makes an important step towards holistic scene reasoning, but does not consider higher-level, stylistic similarity that is often shared across objects in the same scene (e.g., matching furniture set for desk, shelves, chair) which could provide notable insight towards comprehensive scene understanding.

\section{Conclusion}

We have presented \OURSFULL{}, which introduces learned, optimizable part priors for fitting complete part decompositions to objects detected in real-world RGB-D scans.
We learn a latent part space over all object parts, characterized with learned neural implicit functions.
This allows for traversing over the part space at test time to jointly optimize across all parts of an object such that it fits to the observed scan geometry while maintaining consistency with any similar detected objects in the scan.
This results in improved part segmentation as well as completion in noisy, incomplete real-world RGB-D scans.
We hope that this help to open up further avenues towards holistic part-based reasoning in real-world environments. 

\medskip\noindent\textbf{Acknowledgements.}
This work was supported by the  Bavarian State Ministry of Science and the Arts coordinated by the Bavarian Research Institute for Digital Transformation (bidt) and the German Research Foundation (DFG) Grant ``Learning How to Interact with Scenes through Part-Based Understanding.''

%%%%%%%%% REFERENCES
{\small
\bibliographystyle{ieee_fullname}
% \bibliography{egbib}

\begin{thebibliography}{10}\itemsep=-1pt

\bibitem{avetisyan2019scan2cad}
Armen Avetisyan, Manuel Dahnert, Angela Dai, Manolis Savva, Angel~X. Chang, and
  Matthias Nie{\ss}ner.
\newblock Scan2cad: Learning {CAD} model alignment in {RGB-D} scans.
\newblock In {\em {IEEE} Conference on Computer Vision and Pattern Recognition,
  {CVPR} 2019, Long Beach, CA, USA, June 16-20, 2019}, pages 2614--2623, 2019.

\bibitem{besl1992method}
Paul~J Besl and Neil~D McKay.
\newblock Method for registration of 3-d shapes.
\newblock In {\em Sensor fusion IV: control paradigms and data structures},
  volume 1611, pages 586--606. Spie, 1992.

\bibitem{bokhovkin2021towards}
Alexey Bokhovkin, Vladislav Ishimtsev, Emil Bogomolov, Denis Zorin, Alexey
  Artemov, Evgeny Burnaev, and Angela Dai.
\newblock Towards part-based understanding of rgb-d scans.
\newblock In {\em Proceedings of the IEEE/CVF Conference on Computer Vision and
  Pattern Recognition}, pages 7484--7494, 2021.

\bibitem{Matterport3D}
Angel~X. Chang, Angela Dai, Thomas~A. Funkhouser, Maciej Halber, Matthias
  Nie{\ss}ner, Manolis Savva, Shuran Song, Andy Zeng, and Yinda Zhang.
\newblock Matterport3d: Learning from {RGB-D} data in indoor environments.
\newblock In {\em 2017 International Conference on 3D Vision, 3DV 2017,
  Qingdao, China, October 10-12, 2017}, pages 667--676, 2017.

\bibitem{shapenet2015}
Angel~X Chang, Thomas Funkhouser, Leonidas Guibas, Pat Hanrahan, Qixing Huang,
  Zimo Li, Silvio Savarese, Manolis Savva, Shuran Song, Hao Su, et~al.
\newblock Shapenet: An information-rich 3d model repository.
\newblock {\em arXiv preprint arXiv:1512.03012}, 2015.

\bibitem{chen2019imnet}
Zhiqin Chen and Hao Zhang.
\newblock Learning implicit fields for generative shape modeling.
\newblock In {\em 2019 IEEE/CVF Conference on Computer Vision and Pattern
  Recognition (CVPR)}, pages 5932--5941, 2019.

\bibitem{chibane2020ifnet}
Julian Chibane, Thiemo Alldieck, and Gerard Pons-Moll.
\newblock Implicit functions in feature space for 3d shape reconstruction and
  completion.
\newblock pages 6968--6979, 06 2020.

\bibitem{chibane2020ndf}
Julian Chibane, Aymen Mir, and Gerard Pons-Moll.
\newblock Neural unsigned distance fields for implicit function learning.
\newblock In {\em Advances in Neural Information Processing Systems
  ({NeurIPS})}, December 2020.

\bibitem{choi2015robust}
Sungjoon Choi, Qian-Yi Zhou, and Vladlen Koltun.
\newblock Robust reconstruction of indoor scenes.
\newblock In {\em Proceedings of the IEEE Conference on Computer Vision and
  Pattern Recognition}, pages 5556--5565, 2015.

\bibitem{choy20194d}
Christopher~B. Choy, JunYoung Gwak, and Silvio Savarese.
\newblock 4d spatio-temporal convnets: Minkowski convolutional neural networks.
\newblock In {\em {IEEE} Conference on Computer Vision and Pattern Recognition,
  {CVPR} 2019, Long Beach, CA, USA, June 16-20, 2019}, pages 3075--3084, 2019.

\bibitem{dai2017scannet}
Angela Dai, Angel~X. Chang, Manolis Savva, Maciej Halber, Thomas Funkhouser,
  and Matthias Niessner.
\newblock Scannet: Richly-annotated 3d reconstructions of indoor scenes.
\newblock In {\em The IEEE Conference on Computer Vision and Pattern
  Recognition (CVPR)}, July 2017.

\bibitem{dai2020sgnn}
Angela Dai, Christian Diller, and Matthias Nie{\ss}ner.
\newblock Sg-nn: Sparse generative neural networks for self-supervised scene
  completion of rgb-d scans.
\newblock In {\em Proceedings of the IEEE/CVF Conference on Computer Vision and
  Pattern Recognition}, pages 849--858, 2020.

\bibitem{dai20183dmv}
Angela Dai and Matthias Nie{\ss}ner.
\newblock 3dmv: Joint 3d-multi-view prediction for 3d semantic scene
  segmentation.
\newblock In {\em Proceedings of the European Conference on Computer Vision
  (ECCV)}, pages 452--468, 2018.

\bibitem{dai2017bundlefusion}
Angela Dai, Matthias Nie{\ss}ner, Michael Zollh{\"o}fer, Shahram Izadi, and
  Christian Theobalt.
\newblock Bundlefusion: Real-time globally consistent 3d reconstruction using
  on-the-fly surface reintegration.
\newblock {\em ACM Transactions on Graphics (ToG)}, 36(4):1, 2017.

\bibitem{dai2018scancomplete}
Angela Dai, Daniel Ritchie, Martin Bokeloh, Scott Reed, J{\"u}rgen Sturm, and
  Matthias Nie{\ss}ner.
\newblock Scancomplete: Large-scale scene completion and semantic segmentation
  for 3d scans.
\newblock In {\em Proceedings of the IEEE Conference on Computer Vision and
  Pattern Recognition}, pages 4578--4587, 2018.

\bibitem{dai2017shape}
Angela Dai, Charles Ruizhongtai~Qi, and Matthias Nie{\ss}ner.
\newblock Shape completion using 3d-encoder-predictor cnns and shape synthesis.
\newblock In {\em Proceedings of the IEEE Conference on Computer Vision and
  Pattern Recognition}, pages 5868--5877, 2017.

\bibitem{dai2021spsg}
Angela Dai, Yawar Siddiqui, Justus Thies, Julien Valentin, and Matthias
  Nie{\ss}ner.
\newblock Spsg: Self-supervised photometric scene generation from rgb-d scans.
\newblock In {\em Proceedings of the IEEE/CVF Conference on Computer Vision and
  Pattern Recognition}, pages 1747--1756, 2021.

\bibitem{engelmann20203d}
Francis Engelmann, Martin Bokeloh, Alireza Fathi, Bastian Leibe, and Matthias
  Nie{\ss}ner.
\newblock 3d-mpa: Multi-proposal aggregation for 3d semantic instance
  segmentation.
\newblock In {\em Proceedings of the IEEE/CVF Conference on Computer Vision and
  Pattern Recognition}, pages 9031--9040, 2020.

\bibitem{geiger2012we}
Andreas Geiger, Philip Lenz, and Raquel Urtasun.
\newblock Are we ready for autonomous driving? the kitti vision benchmark
  suite.
\newblock In {\em 2012 IEEE conference on computer vision and pattern
  recognition}, pages 3354--3361. IEEE, 2012.

\bibitem{genova2020local}
Kyle Genova, Forrester Cole, Avneesh Sud, Aaron Sarna, and Thomas Funkhouser.
\newblock Local deep implicit functions for 3d shape.
\newblock In {\em Proceedings of the IEEE/CVF Conference on Computer Vision and
  Pattern Recognition}, pages 4857--4866, 2020.

\bibitem{genova2019learning}
Kyle Genova, Forrester Cole, Daniel Vlasic, Aaron Sarna, William~T Freeman, and
  Thomas Funkhouser.
\newblock Learning shape templates with structured implicit functions.
\newblock In {\em Proceedings of the IEEE/CVF International Conference on
  Computer Vision}, pages 7154--7164, 2019.

\bibitem{golovinskiy2009consistent}
Aleksey Golovinskiy and Thomas Funkhouser.
\newblock Consistent segmentation of 3d models.
\newblock {\em Computers \& Graphics}, 33(3):262--269, 2009.

\bibitem{graham20183dsemantic}
Benjamin Graham, Martin Engelcke, and Laurens van~der Maaten.
\newblock 3d semantic segmentation with submanifold sparse convolutional
  networks.
\newblock In {\em 2018 {IEEE} Conference on Computer Vision and Pattern
  Recognition, {CVPR} 2018, Salt Lake City, UT, USA, June 18-22, 2018}, pages
  9224--9232, 2018.

\bibitem{han2020occuseg}
Lei Han, Tian Zheng, Lan Xu, and Lu Fang.
\newblock Occuseg: Occupancy-aware 3d instance segmentation.
\newblock In {\em Proceedings of the IEEE/CVF Conference on Computer Vision and
  Pattern Recognition}, pages 2940--2949, 2020.

\bibitem{hanocka2019meshcnn}
Rana Hanocka, Amir Hertz, Noa Fish, Raja Giryes, Shachar Fleishman, and Daniel
  Cohen-Or.
\newblock Meshcnn: a network with an edge.
\newblock {\em ACM Transactions on Graphics (TOG)}, 38(4):1--12, 2019.

\bibitem{hou20193dsis}
Ji Hou, Angela Dai, and Matthias Nie{\ss}ner.
\newblock 3d-sis: 3d semantic instance segmentation of rgb-d scans.
\newblock In {\em Proceedings of the IEEE Conference on Computer Vision and
  Pattern Recognition}, pages 4421--4430, 2019.

\bibitem{hou2020revealnet}
Ji Hou, Angela Dai, and Matthias Nie{\ss}ner.
\newblock Revealnet: Seeing behind objects in rgb-d scans.
\newblock In {\em Proceedings of the IEEE/CVF Conference on Computer Vision and
  Pattern Recognition}, pages 2098--2107, 2020.

\bibitem{hu2012co}
Ruizhen Hu, Lubin Fan, and Ligang Liu.
\newblock Co-segmentation of 3d shapes via subspace clustering.
\newblock In {\em Computer graphics forum}, volume~31, pages 1703--1713. Wiley
  Online Library, 2012.

\bibitem{huang2011joint}
Qixing Huang, Vladlen Koltun, and Leonidas Guibas.
\newblock Joint shape segmentation with linear programming.
\newblock In {\em Proceedings of the 2011 SIGGRAPH Asia Conference}, pages
  1--12, 2011.

\bibitem{jiang2020pointgroup}
Li Jiang, Hengshuang Zhao, Shaoshuai Shi, Shu Liu, Chi-Wing Fu, and Jiaya Jia.
\newblock Pointgroup: Dual-set point grouping for 3d instance segmentation.
\newblock In {\em Proceedings of the IEEE/CVF Conference on Computer Vision and
  Pattern Recognition}, pages 4867--4876, 2020.

\bibitem{kalogerakis20173d}
Evangelos Kalogerakis, Melinos Averkiou, Subhransu Maji, and Siddhartha
  Chaudhuri.
\newblock 3d shape segmentation with projective convolutional networks.
\newblock In {\em proceedings of the IEEE conference on computer vision and
  pattern recognition}, pages 3779--3788, 2017.

\bibitem{lorensen1987marching}
William~E Lorensen and Harvey~E Cline.
\newblock Marching cubes: A high resolution 3d surface construction algorithm.
\newblock {\em ACM siggraph computer graphics}, 21(4):163--169, 1987.

\bibitem{luo2020learning}
Tiange Luo, Kaichun Mo, Zhiao Huang, Jiarui Xu, Siyu Hu, Liwei Wang, and Hao
  Su.
\newblock Learning to group: A bottom-up framework for 3d part discovery in
  unseen categories.
\newblock {\em arXiv preprint arXiv:2002.06478}, 2020.

\bibitem{mildenhall2020nerf}
Ben Mildenhall, Pratul~P. Srinivasan, Matthew Tancik, Jonathan~T. Barron, Ravi
  Ramamoorthi, and Ren Ng.
\newblock Nerf: Representing scenes as neural radiance fields for view
  synthesis.
\newblock In {\em ECCV}, 2020.

\bibitem{mo2019structurenet}
Kaichun Mo, Paul Guerrero, Li Yi, Hao Su, Peter Wonka, Niloy Mitra, and
  Leonidas~J Guibas.
\newblock Structurenet: Hierarchical graph networks for 3d shape generation.
\newblock {\em arXiv preprint arXiv:1908.00575}, 2019.

\bibitem{mo2019partnet}
Kaichun Mo, Shilin Zhu, Angel~X Chang, Li Yi, Subarna Tripathi, Leonidas~J
  Guibas, and Hao Su.
\newblock Partnet: A large-scale benchmark for fine-grained and hierarchical
  part-level 3d object understanding.
\newblock In {\em Proceedings of the IEEE Conference on Computer Vision and
  Pattern Recognition}, pages 909--918, 2019.

\bibitem{newcombe2011kinectfusion}
Richard~A Newcombe, Shahram Izadi, Otmar Hilliges, David Molyneaux, David Kim,
  Andrew~J Davison, Pushmeet Kohi, Jamie Shotton, Steve Hodges, and Andrew
  Fitzgibbon.
\newblock Kinectfusion: Real-time dense surface mapping and tracking.
\newblock In {\em 2011 10th IEEE international symposium on mixed and augmented
  reality}, pages 127--136. IEEE, 2011.

\bibitem{nie2021rfd}
Yinyu Nie, Ji Hou, Xiaoguang Han, and Matthias Nie{\ss}ner.
\newblock Rfd-net: Point scene understanding by semantic instance
  reconstruction.
\newblock In {\em Proceedings of the IEEE/CVF Conference on Computer Vision and
  Pattern Recognition}, pages 4608--4618, 2021.

\bibitem{niessner2013real}
Matthias Nie{\ss}ner, Michael Zollh{\"o}fer, Shahram Izadi, and Marc
  Stamminger.
\newblock Real-time 3d reconstruction at scale using voxel hashing.
\newblock {\em ACM Transactions on Graphics (ToG)}, 32(6):1--11, 2013.

\bibitem{park2019deepsdf}
Jeong~Joon Park, Peter Florence, Julian Straub, Richard Newcombe, and Steven
  Lovegrove.
\newblock Deepsdf: Learning continuous signed distance functions for shape
  representation.
\newblock In {\em Proceedings of the IEEE Conference on Computer Vision and
  Pattern Recognition}, pages 165--174, 2019.

\bibitem{peng2020convolutional}
Songyou Peng, Michael Niemeyer, Lars Mescheder, Marc Pollefeys, and Andreas
  Geiger.
\newblock Convolutional occupancy networks.
\newblock In {\em Computer Vision--ECCV 2020: 16th European Conference,
  Glasgow, UK, August 23--28, 2020, Proceedings, Part III 16}, pages 523--540.
  Springer, 2020.

\bibitem{qi2020imvotenet}
Charles~R Qi, Xinlei Chen, Or Litany, and Leonidas~J Guibas.
\newblock Imvotenet: Boosting 3d object detection in point clouds with image
  votes.
\newblock In {\em Proceedings of the IEEE/CVF conference on computer vision and
  pattern recognition}, pages 4404--4413, 2020.

\bibitem{qi2019deep}
Charles~R Qi, Or Litany, Kaiming He, and Leonidas~J Guibas.
\newblock Deep hough voting for 3d object detection in point clouds.
\newblock In {\em Proceedings of the IEEE International Conference on Computer
  Vision}, pages 9277--9286, 2019.

\bibitem{qi2017pointnet}
Charles~Ruizhongtai Qi, Hao Su, Kaichun Mo, and Leonidas~J. Guibas.
\newblock Pointnet: Deep learning on point sets for 3d classification and
  segmentation.
\newblock In {\em 2017 {IEEE} Conference on Computer Vision and Pattern
  Recognition, {CVPR} 2017, Honolulu, HI, USA, July 21-26, 2017}, pages 77--85,
  2017.

\bibitem{rusinkiewicz2001efficient}
Szymon Rusinkiewicz and Marc Levoy.
\newblock Efficient variants of the icp algorithm.
\newblock In {\em Proceedings third international conference on 3-D digital
  imaging and modeling}, pages 145--152. IEEE, 2001.

\bibitem{sidi2011unsupervised}
Oana Sidi, Oliver van Kaick, Yanir Kleiman, Hao Zhang, and Daniel Cohen-Or.
\newblock Unsupervised co-segmentation of a set of shapes via descriptor-space
  spectral clustering.
\newblock In {\em Proceedings of the 2011 SIGGRAPH Asia Conference}, pages
  1--10, 2011.

\bibitem{sitzmann2019siren}
Vincent Sitzmann, Julien~N.P. Martel, Alexander~W. Bergman, David~B. Lindell,
  and Gordon Wetzstein.
\newblock Implicit neural representations with periodic activation functions.
\newblock In {\em Proc. NeurIPS}, 2020.

\bibitem{song2017semantic}
Shuran Song, Fisher Yu, Andy Zeng, Angel~X Chang, Manolis Savva, and Thomas
  Funkhouser.
\newblock Semantic scene completion from a single depth image.
\newblock In {\em Proceedings of the IEEE Conference on Computer Vision and
  Pattern Recognition}, pages 1746--1754, 2017.

\bibitem{tung2019learning}
Hsiao-Yu~Fish Tung, Ricson Cheng, and Katerina Fragkiadaki.
\newblock Learning spatial common sense with geometry-aware recurrent networks.
\newblock In {\em Proceedings of the IEEE/CVF Conference on Computer Vision and
  Pattern Recognition}, pages 2595--2603, 2019.

\bibitem{van2013co}
Oliver Van~Kaick, Kai Xu, Hao Zhang, Yanzhen Wang, Shuyang Sun, Ariel Shamir,
  and Daniel Cohen-Or.
\newblock Co-hierarchical analysis of shape structures.
\newblock {\em ACM Transactions on Graphics (TOG)}, 32(4):1--10, 2013.

\bibitem{wang2011symmetry}
Yanzhen Wang, Kai Xu, Jun Li, Hao Zhang, Ariel Shamir, Ligang Liu, Zhiquan
  Cheng, and Yueshan Xiong.
\newblock Symmetry hierarchy of man-made objects.
\newblock In {\em Computer graphics forum}, volume~30, pages 287--296. Wiley
  Online Library, 2011.

\bibitem{whelan2015elasticfusion}
Thomas Whelan, Stefan Leutenegger, R Salas-Moreno, Ben Glocker, and Andrew
  Davison.
\newblock Elasticfusion: Dense slam without a pose graph.
\newblock Robotics: Science and Systems, 2015.

\bibitem{wu2020pq}
Rundi Wu, Yixin Zhuang, Kai Xu, Hao Zhang, and Baoquan Chen.
\newblock Pq-net: A generative part seq2seq network for 3d shapes.
\newblock In {\em Proceedings of the IEEE/CVF Conference on Computer Vision and
  Pattern Recognition}, pages 829--838, 2020.

\bibitem{wu20153d}
Zhirong Wu, Shuran Song, Aditya Khosla, Fisher Yu, Linguang Zhang, Xiaoou Tang,
  and Jianxiong Xiao.
\newblock 3d shapenets: {A} deep representation for volumetric shapes.
\newblock In {\em {IEEE} Conference on Computer Vision and Pattern Recognition,
  {CVPR} 2015, Boston, MA, USA, June 7-12, 2015}, pages 1912--1920, 2015.

\bibitem{xie2020mlcvnet}
Qian Xie, Yu-Kun Lai, Jing Wu, Zhoutao Wang, Yiming Zhang, Kai Xu, and Jun
  Wang.
\newblock Mlcvnet: Multi-level context votenet for 3d object detection.
\newblock In {\em Proceedings of the IEEE/CVF Conference on Computer Vision and
  Pattern Recognition}, pages 10447--10456, 2020.

\bibitem{yi2017learning}
Li Yi, Leonidas Guibas, Aaron Hertzmann, Vladimir~G Kim, Hao Su, and Ersin
  Yumer.
\newblock Learning hierarchical shape segmentation and labeling from online
  repositories.
\newblock {\em arXiv preprint arXiv:1705.01661}, 2017.

\bibitem{yi2016scalable}
Li Yi, Vladimir~G Kim, Duygu Ceylan, I-Chao Shen, Mengyan Yan, Hao Su, Cewu Lu,
  Qixing Huang, Alla Sheffer, and Leonidas Guibas.
\newblock A scalable active framework for region annotation in 3d shape
  collections.
\newblock {\em ACM Transactions on Graphics (ToG)}, 35(6):1--12, 2016.

\bibitem{yi2019gspn}
Li Yi, Wang Zhao, He Wang, Minhyuk Sung, and Leonidas~J Guibas.
\newblock Gspn: Generative shape proposal network for 3d instance segmentation
  in point cloud.
\newblock In {\em Proceedings of the IEEE/CVF Conference on Computer Vision and
  Pattern Recognition}, pages 3947--3956, 2019.

\end{thebibliography}

}

\clearpage
\appendix
\begin{table*}[bp]
\centering
\resizebox{\textwidth}{!}{
\begin{tabular}{l|ccccccc|ccccccc|ccccc|}
& \multicolumn{19}{c}{Chamfer Distance ($\downarrow$) -- Accuracy} \\
\toprule
      & \multicolumn{7}{c|}{chair} & \multicolumn{7}{c|}{table} & \multicolumn{5}{c|}{cabinet} \\
\toprule
    Method & left arm & right arm & back & seat & reg. leg & star leg & surf. base & central supp. & drawer & leg & pedestal & shelf & surface & side panel & door & shelf & frame & base & countertop \\
\midrule
    SG-NN\cite{dai2020sgnn} + MLCVNet\cite{xie2020mlcvnet} + PointGroup\cite{jiang2020pointgroup} & 0.081 & 0.075 & 0.045 & 0.012 & 0.022 & 0.099 & 0.200 & 0.164 & 0.204 & \textbf{0.044} & 0.245 & 0.174 & 0.024 & 0.204 & 0.150 & 0.124 & \textbf{0.010} & 0.376 & 0.344 \\
    MLCVNet\cite{xie2020mlcvnet} + StructureNet\cite{mo2019structurenet} & 0.041 & 0.036 & 0.005 & 0.008 & \textbf{0.020} & 0.100 & 0.223 & 0.123 & \textbf{0.021} & 0.105 & 0.167 & \textbf{0.041} & 0.022 & 0.169 & 0.104 & \textbf{0.033} & 0.028 & 0.268 & 0.344 \\
    Bokhovkin et al.\cite{bokhovkin2021towards} & 0.039 & 0.039 & 0.008 & 0.008 & 0.057 & 0.074 & 0.110 & 0.044 & 0.094 & 0.141 & 0.167 & 0.121 & 0.024 & 0.175 & 0.083 & 0.066 & 0.032 & 0.210 & \textbf{0.229} \\
\midrule
    {\bf Ours} & \textbf{0.015} & \textbf{0.015} & \textbf{0.002} & \textbf{0.003} & 0.026 & \textbf{0.058} & \textbf{0.081} & \textbf{0.042} & 0.143 & 0.077 & \textbf{0.101} & 0.146 & \textbf{0.006} & \textbf{0.164} & \textbf{0.042} & 0.078 & 0.017 & \textbf{0.207} & 0.306 \\
\bottomrule
\end{tabular}
}
\caption{Per-part evaluation of semantic part completion for 'chair', 'table', and 'cabinet' categories on Scan2CAD~\cite{avetisyan2019scan2cad} in comparison to state-of-the-art part segmentation \cite{jiang2020pointgroup,mo2019structurenet} and semantic part completion \cite{bokhovkin2021towards}.}
\label{tab:supp_part_acc_comparison_1}
\end{table*}

\begin{table*}[bp]
\centering
\resizebox{\textwidth}{!}{
\begin{tabular}{l|cccc|cccc|ccccc|cc||}
& \multicolumn{15}{c}{Chamfer Distance ($\downarrow$) -- Accuracy}\\
\toprule
      & \multicolumn{4}{c|}{bookshelf} & \multicolumn{4}{c|}{bed} & \multicolumn{5}{c|}{bin} & & \\
\toprule
    Method & door & shelf & frame & base & frame & side surf. & sleep area & headboard & base & bottom & box & cover & frame & class avg. & inst. avg. \\
\midrule
    SG-NN\cite{dai2020sgnn} + MLCVNet\cite{xie2020mlcvnet} + PointGroup\cite{jiang2020pointgroup} & 0.097 & 0.245 & \textbf{0.005} & 1.298 & 0.059 & 0.776 & \textbf{0.009} & 0.890 & 0.191 & 0.072 & \textbf{0.001} & 0.133 & 0.049 & 0.201 & 0.077 \\
    MLCVNet\cite{xie2020mlcvnet} + StructureNet\cite{mo2019structurenet} & 0.070 & 0.186 & 0.045 & 1.161 & 0.081 & 0.776 & 0.047 & 1.375 & 0.191 & 0.044 & 0.003 & 0.126 & 0.049 & 0.188 & 0.055 \\
    Bokhovkin et al.\cite{bokhovkin2021towards} & 0.041 & 0.137 & 0.090 & \textbf{0.858} & 0.073 & 0.485 & 0.127 & 0.508 & 0.131 & 0.038 & 0.004 & 0.042 & 0.041 & 0.134 & 0.054 \\
\midrule
    {\bf Ours} & \textbf{0.037} & \textbf{0.079} & 0.068 & 1.298 & \textbf{0.020} & \textbf{0.290} & 0.051 & \textbf{0.365} & \textbf{0.111} & \textbf{0.020} & 0.002 & \textbf{0.012} & \textbf{0.037} & \textbf{0.123} & \textbf{0.033} \\
\bottomrule
\end{tabular}
}
\caption{Per-part evaluation of semantic part completion for 'bookshelf', 'bed', and 'bin' categories on Scan2CAD~\cite{avetisyan2019scan2cad} in comparison to state-of-the-art part segmentation \cite{jiang2020pointgroup,mo2019structurenet} and semantic part completion \cite{bokhovkin2021towards}.}
\label{tab:supp_part_acc_comparison_2}
\end{table*}

\begin{table*}[bp]
\centering
\resizebox{\textwidth}{!}{
\begin{tabular}{l|ccccccc|ccccccc|ccccc|}
& \multicolumn{19}{c}{Chamfer Distance ($\downarrow$) -- Completion} \\
\toprule
      & \multicolumn{7}{c|}{chair} & \multicolumn{7}{c|}{table} & \multicolumn{5}{c|}{cabinet} \\
\toprule
    Method & left arm & right arm & back & seat & reg. leg & star leg & surf. base & central supp. & drawer & leg & pedestal & shelf & surface & side panel & door & shelf & frame & base & countertop \\
\midrule
    SG-NN\cite{dai2020sgnn} + MLCVNet\cite{xie2020mlcvnet} + PointGroup\cite{jiang2020pointgroup} & 0.096 & 0.086 & 0.041 & 0.014 & 0.054 & 0.063 & 0.161 & 0.152 & 0.137 & \textbf{0.092} & 0.305 & 0.151 & 0.050 & 0.203 & 0.081 & 0.140 & \textbf{0.043} & 0.384 & 0.156 \\
    MLCVNet\cite{xie2020mlcvnet} + StructureNet\cite{mo2019structurenet} & 0.051 & 0.049 & 0.008 & 0.008 & 0.035 & 0.049 & 0.126 & 0.062 & 0.194 & 0.158 & \textbf{0.148} & \textbf{0.102} & 0.037 & 0.246 & 0.104 & \textbf{0.085} & 0.061 & 0.298 & 0.156 \\
    Bokhovkin et al.\cite{bokhovkin2021towards} & 0.046 & 0.049 & 0.012 & 0.011 & 0.056 & 0.057 & 0.139 & 0.078 & 0.149 & 0.165 & \textbf{0.148} & 0.106 & 0.045 & \textbf{0.181} & 0.094 & 0.111 & 0.062 & \textbf{0.219} & \textbf{0.111} \\
\midrule
    {\bf Ours} & \textbf{0.043} & \textbf{0.040} & \textbf{0.006} & \textbf{0.005} & \textbf{0.032} & \textbf{0.031} & \textbf{0.120} & \textbf{0.044} & \textbf{0.135} & 0.118 & \textbf{0.148} & 0.116 & \textbf{0.026} & \textbf{0.181} & \textbf{0.071} & 0.159 & 0.064 & 0.266 & 0.152 \\
\bottomrule
\end{tabular}
}
\caption{Per-part evaluation of part segmentation for 'chair', 'table', and 'cabinet' categories on Scan2CAD~\cite{avetisyan2019scan2cad} in comparison to state-of-the-art part segmentation \cite{jiang2020pointgroup,mo2019structurenet} and semantic part completion \cite{bokhovkin2021towards}.}
\label{tab:supp_part_comp_comparison_1}
\end{table*}

\begin{table*}[bp]
\centering
\resizebox{\textwidth}{!}{
\begin{tabular}{l|cccc|cccc|ccccc|cc||}
& \multicolumn{15}{c}{Chamfer Distance ($\downarrow$) -- Completion}\\
\toprule
      & \multicolumn{4}{c|}{bookshelf} & \multicolumn{4}{c|}{bed} & \multicolumn{5}{c|}{bin} & & \\
\toprule
    Method & door & shelf & frame & base & frame & side surf. & sleep area & headboard & base & bottom & box & cover & frame & class avg. & inst. avg. \\
\midrule
    SG-NN\cite{dai2020sgnn} + MLCVNet\cite{xie2020mlcvnet} + PointGroup\cite{jiang2020pointgroup} & 0.348 & 0.143 & \textbf{0.020} & 1.066 & 0.072 & 0.567 & 0.043 & 0.999 & 0.284 & 0.077 & \textbf{0.004} & 0.093 & 0.038 & 0.193 & 0.084 \\
    MLCVNet\cite{xie2020mlcvnet} + StructureNet\cite{mo2019structurenet} & \textbf{0.173} & 0.131 & 0.071 & 0.942 & 0.077 & 0.567 & 0.057 & 1.122 & 0.284 & 0.052 & 0.007 & 0.096 & 0.038 & 0.175 & 0.062 \\
    Bokhovkin et al.\cite{bokhovkin2021towards} & 0.361 & \textbf{0.095} & 0.096 & \textbf{0.754} & 0.068 & 0.505 & 0.109 & 0.507 & \textbf{0.153} & 0.050 & 0.008 & 0.045 & 0.036 & 0.144 & 0.056 \\
\midrule
    {\bf Ours} & 0.520 & 0.130 & 0.119 & 1.066 & \textbf{0.034} & \textbf{0.221} & \textbf{0.038} & \textbf{0.365} & 0.167 & \textbf{0.029} & \textbf{0.004} & \textbf{0.020} & \textbf{0.033} & \textbf{0.140} & \textbf{0.043} \\
\bottomrule
\end{tabular}
}
\caption{Per-part evaluation of part segmentation for 'bookshelf', 'bed', and 'bin' categories on Scan2CAD~\cite{avetisyan2019scan2cad} in comparison to state-of-the-art part segmentation \cite{jiang2020pointgroup,mo2019structurenet} and semantic part completion \cite{bokhovkin2021towards}.}
\label{tab:supp_part_comp_comparison_2}
\end{table*}

In this supplemental material, we provide additional qualitative results in Section~\ref{sec:supp_addres}, evaluation details in Section~\ref{sec:supp_evaldetails}, additional per-part evaluation in Section~\ref{sec:supp_perpart}, additional ablation discussion with qualitative visualizations for ablations in Section~\ref{sec:supp_ablation}, visualization of part interpolations through learned latent part spaces in Section~\ref{sec:supp_interpolation}, part specifications per category in Section~\ref{sec:supp_parts}, and implementation details with the description of the network architecture in Section~\ref{sec:supp_implementation}.

\section{Additional qualitative analysis}
\label{sec:supp_addres}

\begin{figure*}
\begin{center}
    \centering
    \includegraphics[width=0.70\textwidth]{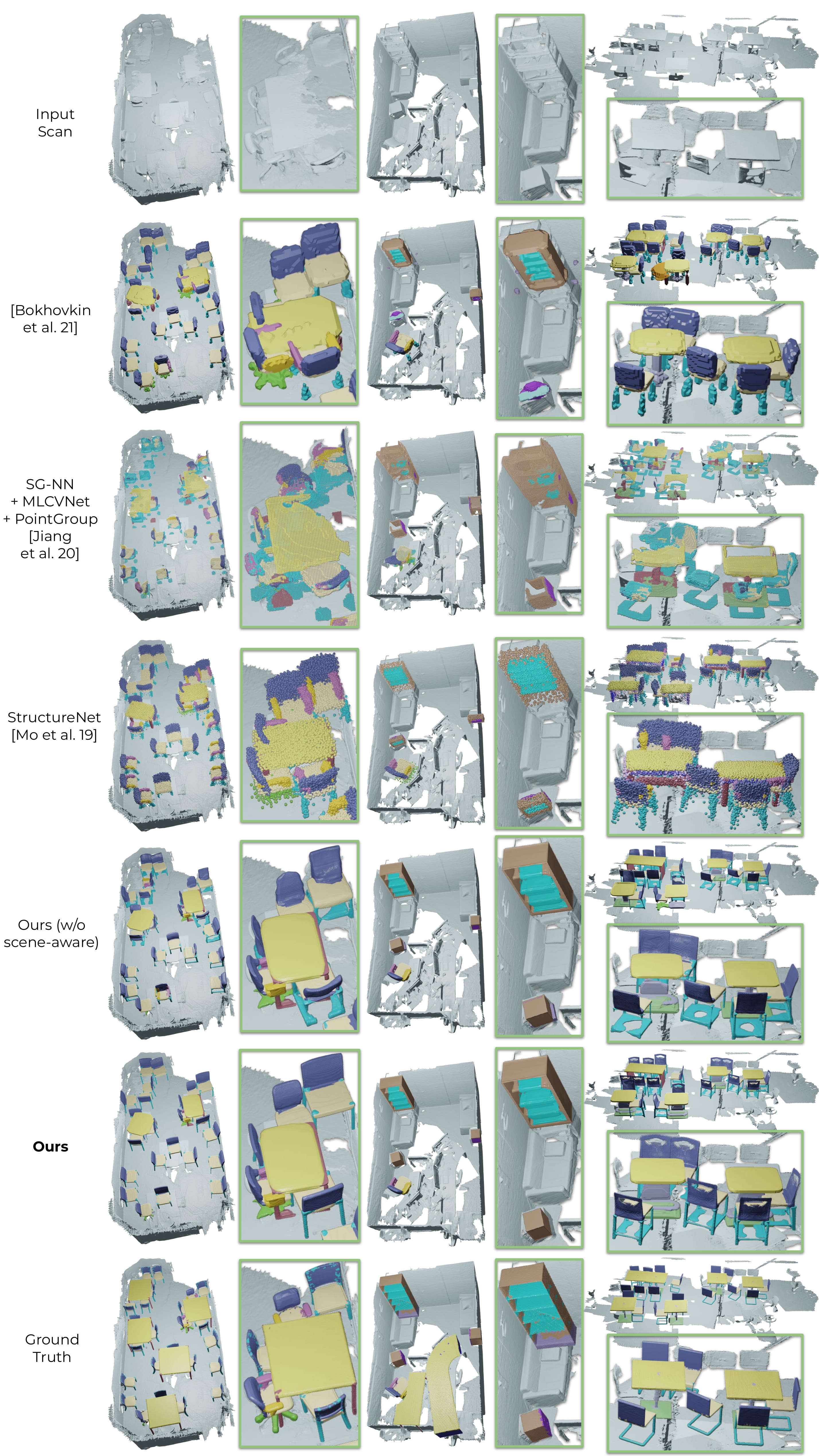}
    \vspace{-0.2cm}
    \caption{Additional qualitative comparison of \OURS{} with point \cite{jiang2020pointgroup,mo2019structurenet} and voxel-based \cite{bokhovkin2021towards} state of the art on ScanNet scans with Scan2CAD+PartNet ground truth.}
    \label{fig:comparison_1}
\end{center}
\end{figure*}

\begin{figure*}
\begin{center}
    \centering
    \includegraphics[width=0.72\textwidth]{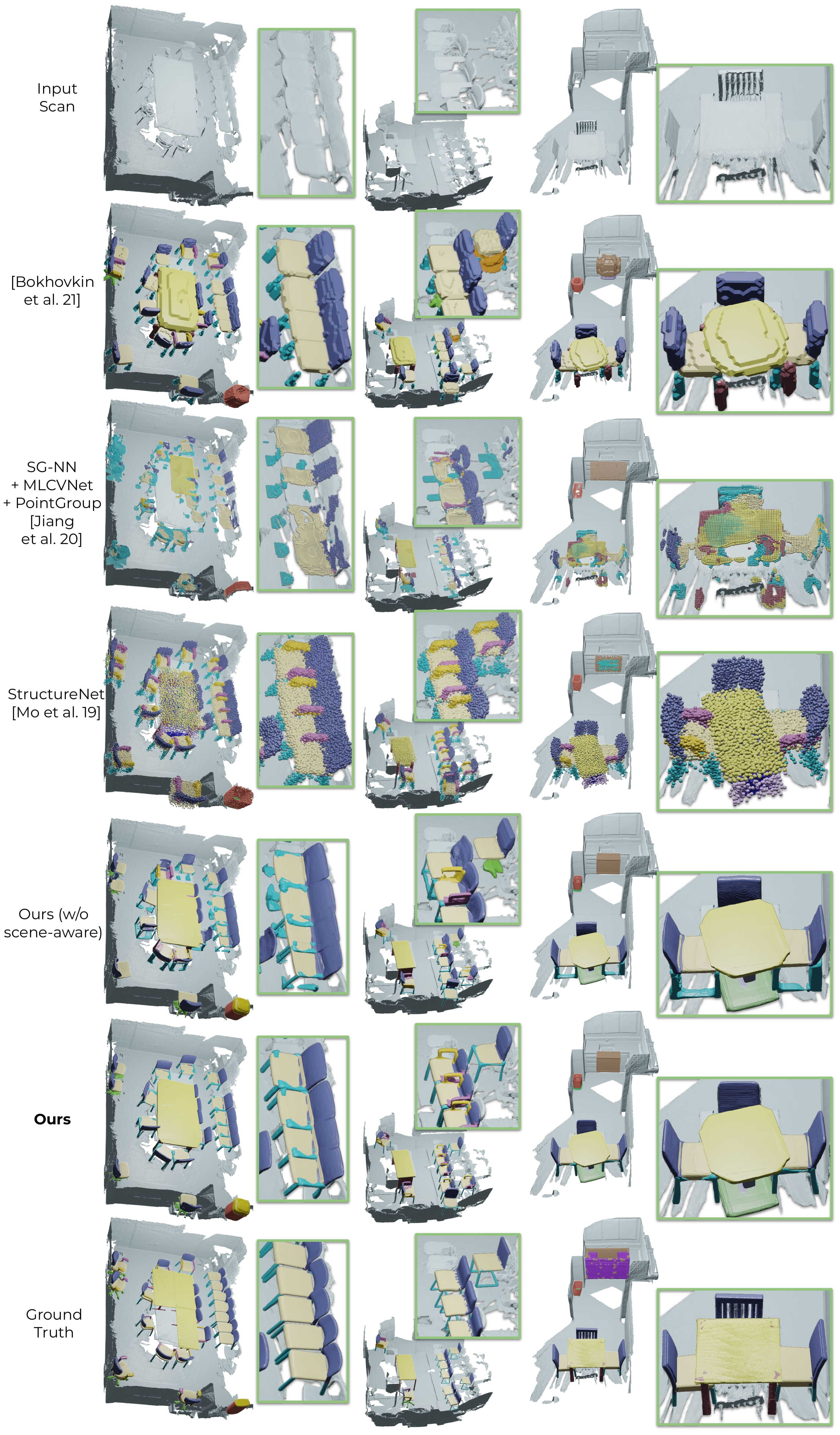}
    \vspace{-0.2cm}
    \caption{Additional qualitative comparison of \OURS{} with point \cite{jiang2020pointgroup,mo2019structurenet} and voxel-based \cite{bokhovkin2021towards} state of the art on ScanNet scans with Scan2CAD+PartNet ground truth.}
    \label{fig:comparison_2}
\end{center}
\end{figure*}

\begin{figure*}
\begin{center}
    \centering
    \vspace{0.3cm}
    \includegraphics[width=0.95\textwidth]{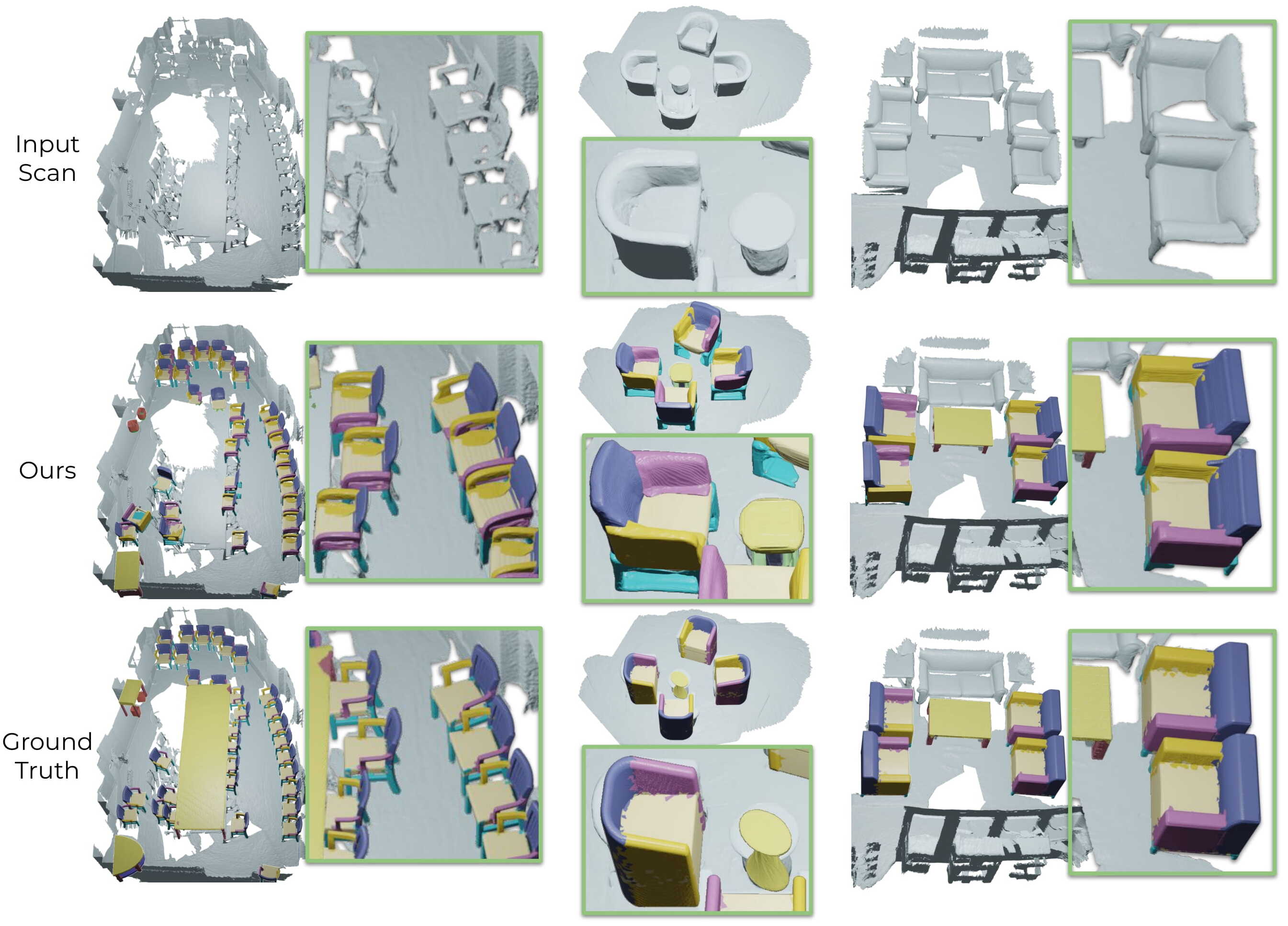}
    \vspace{0.0cm}
    \caption{Additional qualitative results on ScanNet\cite{dai2017scannet} with Scan2CAD\cite{avetisyan2019scan2cad} and PartNet\cite{mo2019partnet} targets, showing our consistent, complete part decompositions.}
    \label{fig:scenes}
    \vspace{0.3cm}
\end{center}%
\end{figure*}

In Figures~\ref{fig:comparison_1}, \ref{fig:comparison_2} and \ref{fig:scenes}, we show additional qualitative results and comparison of \OURS{} to baseline methods. We can see that StructureNet~\cite{mo2019structurenet} and Bokhovkin et al.~\cite{bokhovkin2021towards} often produce incomplete and inaccurate shapes, can include inconsistent parts (e.g. predicting only one chair arm or predicting two types of legs for one chair). The advanced point cloud segmentation method PointGroup~\cite{jiang2020pointgroup} is able to predict consistent part types for shapes but produces fairly noisy geometry for these parts. In addition, when comparing to baselines and \OURS{} without applying scene-aware constraints, we can clearly see a large amount of diversity within shapes that should be similar or identical within one scan. 
%Moreover, it is likely for \OURS{} without applying scene-aware constraints to have parts with bleeding geometry towards floor or walls (best seen for chair legs).

\begin{figure*}
\begin{center}
    \centering
    \includegraphics[width=0.85\textwidth]{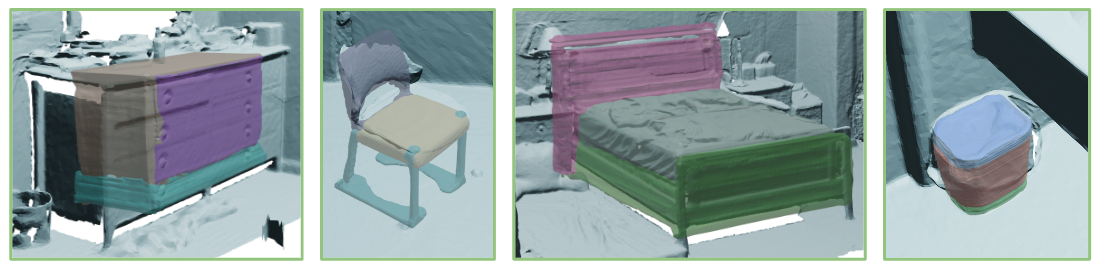}
    \vspace{-0.2cm}
    \caption{Common failure cases for different shape categories produced by NPPs.}
    \label{fig:failure}
\end{center}
\end{figure*}

Several categories can be more challenging, due to more often appearing with clutter (e.g., tables often have objects on top vs chairs or trash cans); our learned manifolds help to regularize this during TTO.
In Fig.~\ref{fig:failure}, we show common failure cases that \OURS{} produces for different shape categories. 
Parts with little geometric distinction (e.g., cabinet frame vs drawer often both lie on a flat plane) can be more difficult to optimize, due to more challenging segmentation. 
The most left case with the cabinet shows erroneous detection (too small), along with an excess wrong part prediction of a cabinet base on the bottom. Missing scanned legs of the chair result in the incorrect type of reconstructed chair legs, while the 4-spoke swivel chair is ground truth. Dense segmentation of real-world scans is often significantly challenging, tending to segment parts of the floor as chair legs, parts of the wall as a bed headboard, or treating a full trash bin as a bin with a top cover part.

% Comparison

% Comparison

% Comparison
% General scenes

% Failure cases

\section{Evaluation details}
\label{sec:supp_evaldetails}

For semantic part completion and part segmentation evaluation, we sample $10,000$ points per part from predicted and ground-truth mesh surfaces (within the corresponding MLCVNet bounding box) and transform them to the ScanNet coordinate space. The Chamfer Distance metric is evaluated for every pair of semantically matching parts between predicted and ground-truth meshes. In case there exists a part in a ground-truth mesh that is missing for a predicted mesh or vice versa, we use the center of the mesh as a missing part. After each part is evaluated, we average scores to obtain the final score for a full object. 

For segmentation evaluation, we project labels from sampled points onto ScanNet mesh vertices to obtain the set of points not depending on a method. Here, to compute segmentation Chamfer Distance for an object, the predicted and ground-truth projected labeled points are used to get per-part scores and then averaged. For one part IoU evaluation, the corresponding projected points are marked as ones and the rest as zeros, intersection and union scores are computed using these 0-1 sets. We similarly use $10,000$ sampled points per part for these metrics.

\section{Per-part evaluation}
\label{sec:supp_perpart}

We provide per-part semantic part completion and part segmentation in Tab.~\ref{tab:supp_part_acc_comparison_1},~\ref{tab:supp_part_acc_comparison_2},~\ref{tab:supp_part_comp_comparison_1}, and~\ref{tab:supp_part_comp_comparison_2}. Our learned part manifolds enable more robust, accurate geometry reconstruction also for individual parts.

\section{Additional ablation discussion}
\label{sec:supp_ablation}

In Tab.~\ref{tab:ablation_scene} we compare results without Scene Consistency constraints to ours only for the instances affected by these constraints. Evaluated only on $\sim 66\%$ of instances and $\sim 57\%$ of corresponding ScanNet scenes, NPPs outperform the results without Scene Consistency constraints by a greater gap compared to Tab.~\ref{tab:ablations}. 

\begin{figure*}
\begin{center}
    \centering
    \includegraphics[width=0.73\textwidth]{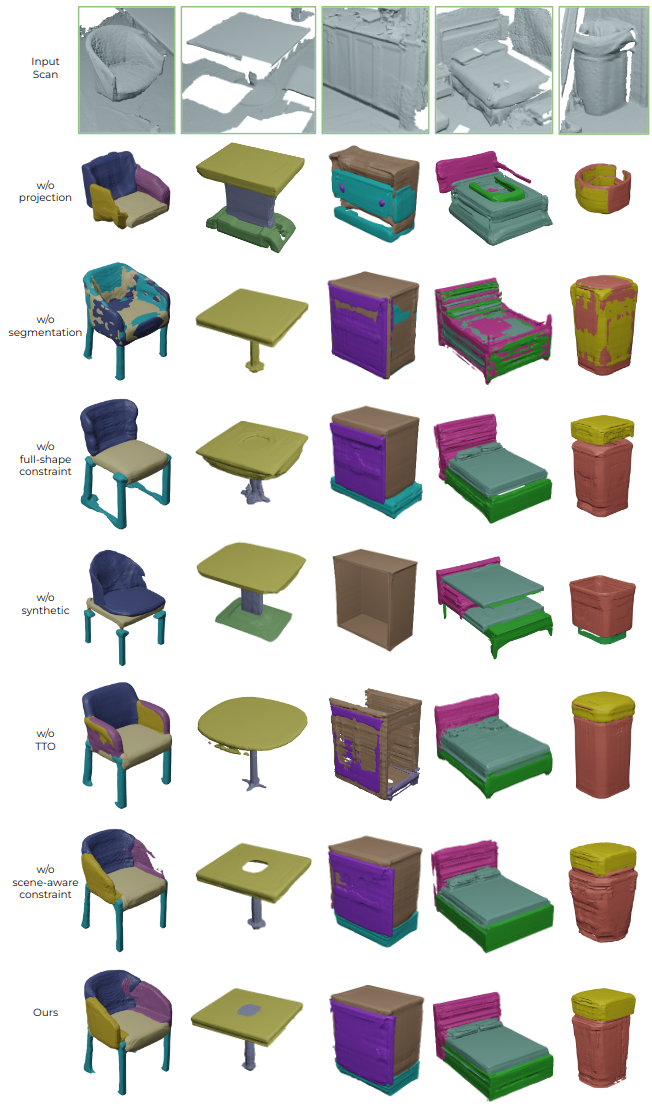}
    \vspace{-0.2cm}
    \caption{Qualitative ablation results on ScanNet\cite{dai2017scannet} with Scan2CAD\cite{avetisyan2019scan2cad} and PartNet\cite{mo2019partnet} targets, showing the importance of every design choice in our method.}
    \label{fig:ablation}
\end{center}
\end{figure*}

In Fig.~\ref{fig:ablation} we show the qualitative ablation results corresponding to Tab.~\ref{tab:ablations}. Compared to quantitative results of scene-aware constraint ablation, we see a more noticeable effect in qualitative effect, with much more consistent part decompositions for similar objects in a scene, even when seen under fairly different partial views (i.e., matching left and right chair arms, consistent joint of the table surface and the table stand). 

Without latent projection, arbitrary initializations for TTO often land outside the basin of convergence (i.e., starting with too discrepant parts for the trash bin and the cabinet), resulting in poorer performance, as low-level geometric constraints may be ambiguous for resolving large structural differences. TTO then improves significantly the fitting accuracy enabling the prediction of geometry that lies outside of the learned manifold (i.e., round table surface, missing pillows on the bed). By leveraging TTO and projection initialization, we can achieve the best representation of the input scan as its part decomposition.

We evaluate the effect of synthetic pre-training of the part segmentation in Tab.~\ref{tab:ablations} (\emph{w/o Synthetic Pretrain}). The additional quantity and diversity of data help to avoid overfitting to more limited real data (i.e., very poor results for 'bed' and 'trashcan' categories due to limited real-world data).

The full-shape constraint helps to maintain consistency between the global shape and the optimized parts during test-time optimization (i.e. not connected box and cover parts for the trash bin, inconsistent joints for the chair and the cabinet). Dense segmentation guides the TTO optimization constraints and prevents self-intersections between parts.

% Ablation

\begin{table*}[h]
\centering
\resizebox{\textwidth}{!}{
\begin{tabular}{l|cccccc|cc||cccccc|cc}
& \multicolumn{8}{c||}{Chamfer Distance ($\downarrow$) -- Accuracy} & \multicolumn{8}{c}{Chamfer Distance ($\downarrow$) -- Completion}\\
\toprule
    Method & chair & table & cab. & bkshlf & bed & bin & class avg & inst avg & chair & table & cab. & bkshlf & bed & bin & class avg & inst avg \\
\midrule
    \# scenes & 126 / 189 & 30 / 127 & 14 / 94 & 13 / 39 & 14 / 47 & 10 / 74 & 164 / 285 & 164 / 285 & 126 / 189 & 30 / 127 & 14 / 94 & 13 / 39 & 14 / 47 & 10 / 74 & 164 / 285 & 164 / 285 \\
    \# instances & 764 / 904 & 81 / 190 & 32 / 140 & 24 / 54 & 28 / 61 & 18 / 85 & 947 / 1434 & 947 / 1434 & 764 / 904 & 81 / 190 & 32 / 140 & 24 / 54 & 28 / 61 & 18 / 85 & 947 / 1434 & 947 / 1434 \\
\midrule
    w/o Scene Consistency & 0.016 & 0.055 & \textbf{0.054} & 0.174 & 0.129 & 0.043 & 0.078 & 0.028 & 0.020 & 0.065 & \textbf{0.112} & \textbf{0.225} & 0.137 & 0.042 & 0.100 & \textbf{0.033} \\
    {\bf Ours} & \textbf{0.011} & \textbf{0.047} & 0.057 & \textbf{0.163} & \textbf{0.101} & \textbf{0.036} & \textbf{0.069} & \textbf{0.023} & \textbf{0.019} & \textbf{0.063} & 0.119 & 0.257 & \textbf{0.099} & \textbf{0.035} & \textbf{0.098} & \textbf{0.033} \\
\bottomrule
\end{tabular}
}
\vspace{-0.3cm}
\caption{Ablation study evaluating semantic part completion on Scan2CAD~\cite{avetisyan2019scan2cad} including only the instances affected with Scene Consistency constraints. We also provide information of how many shape instances and scenes are affected with scene-aware constraints for each category.}
\label{tab:ablation_scene}
\end{table*}

\section{Interpolation properties of learned latent part spaces}
\label{sec:supp_interpolation}

\begin{figure*}
\begin{center}
    \centering
    \includegraphics[width=0.9\textwidth]{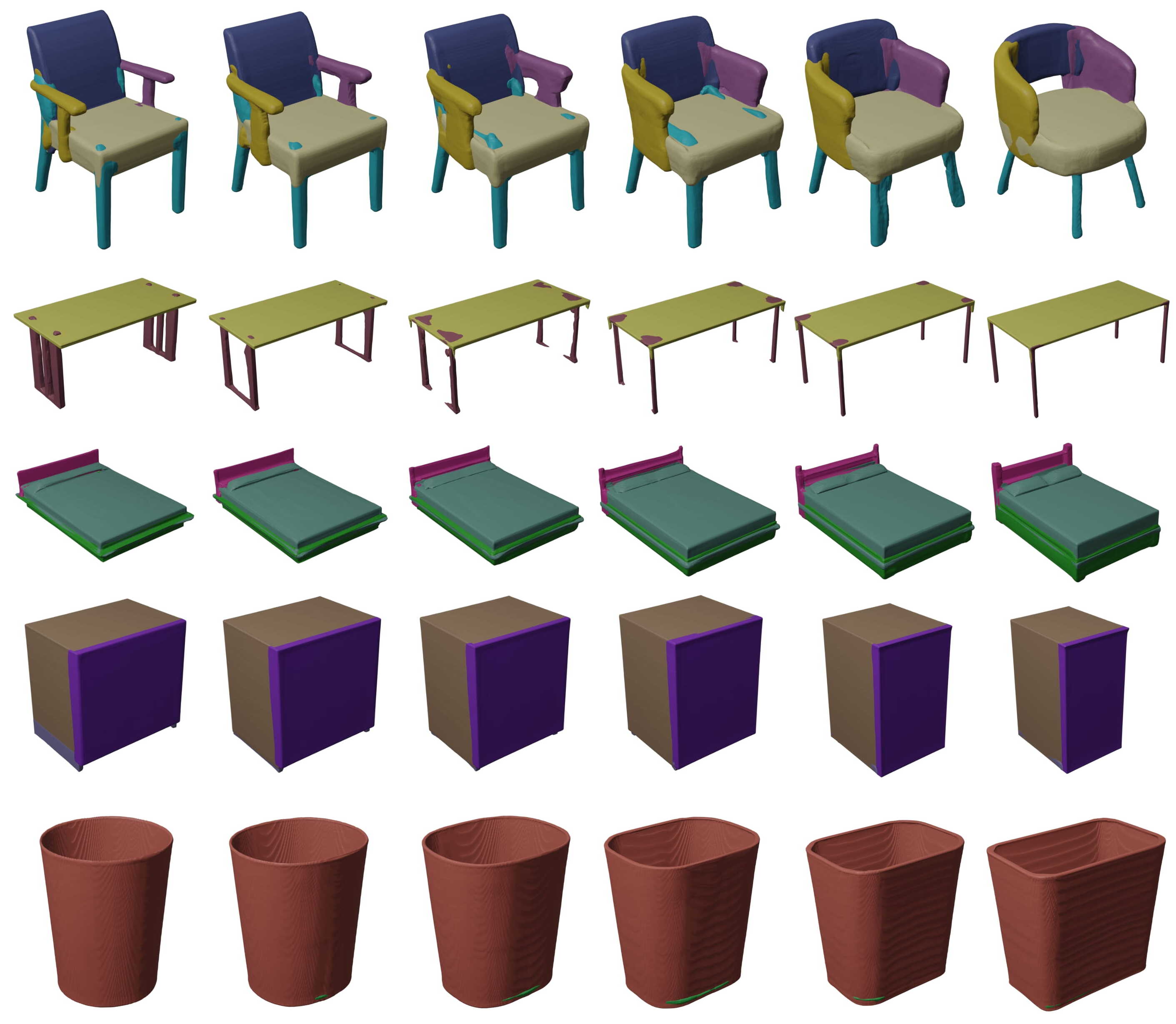}
    \caption{Part interpolations through our learned latent part spaces for different shape classes.}
    \label{fig:interpolation}
\end{center}
\end{figure*}

In Figure~\ref{fig:interpolation}, we show the interpolation capabilities of the part latent spaces that we use in \OURS{} to traverse during test-time optimization. 
Although each part space has been learned individually, their interpolations can produce consistent shapes. 

% Interpolation

\section{Part types per category}
\label{sec:supp_parts}

\begin{figure*}
\begin{center}
    \centering
    \includegraphics[width=0.9\textwidth]{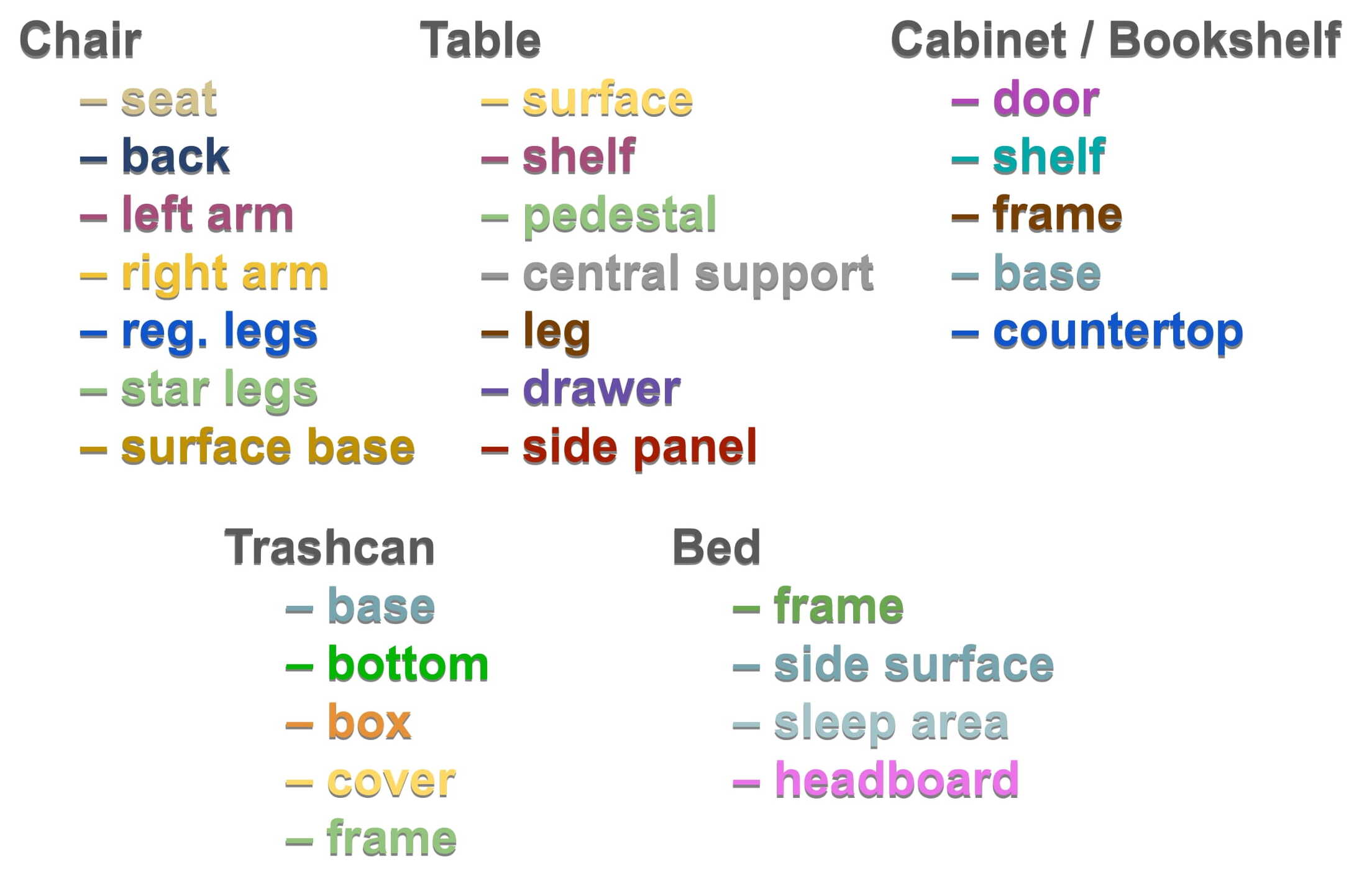}
    \caption{Part specifications per category for the parts used in our approach. Note that 'cabinet' and 'bookshelf' have the same set of parts.}
    \label{fig:parts}
\end{center}
\end{figure*}

In Figure~\ref{fig:parts} we present the shape categories and the corresponding parts that we use in our framework. There are 6 shape categories and 28 part types in total.

% Parts

\section{Implementation details}
\label{sec:supp_implementation}

We provide further implementation details; note that parameters reported here are for the `chair' category, and other category parameter differences are specified in Table~\ref{tab:hyper}.

\subsection{Pretrain decoders}
We first train our latent part and shape spaces on the synthetic PartNet~\cite{mo2019partnet} dataset. This corresponds to the tasks `Train decoder (shape)', `Train decoder (parts)' in Table~\ref{tab:hyper}.
The part and shape decoders are all MLPs composed of 8 linear layers of 512 dimensions each, using ReLU nonlinearities with a final tanh for SDF output. The detailed architecture is shown in Tables~\ref{tab:arch_shape_sdf}, \ref{tab:arch_parts_sdf}.
To train the shape decoder, we use an Adam optimizer with a batch size of 24 and learning rate of 5e-5 (`lr' in the Table~\ref{tab:hyper}) for network weights (with a factor 0.5 (`lr factor') and decay interval of 500 epochs (`lr decay int.')) and 1e-4 for latent parameters (with a factor 0.5 and decay interval of 500 epochs), and train for 2000 epochs. For the part decoder, we extend every part latent with one-hot encoded part type and train the part decoder using an Adam optimizer with a batch size of 48 and learning rate of 5e-5 for network weights (with a decay factor of 0.5 and decay interval of 400 epochs) and 1e-4 for latent parameters (with a decay factor 0.5 and decay interval of 400 epochs), and train for 2000 epochs. 

\subsection{Pre-training for latent projection and part segmentation}
We then train the projection mapping into the learned part and shape spaces as well as the part segmentation.
This part corresponds to the task `Train projection' in Table~\ref{tab:hyper}.
Our model is pre-trained on synthetic PartNet data using virtually scanned incomplete inputs to take advantage of a large amount of synthetic data. We use an Adam optimizer with batch size 64 and learning rate 1e-3 (`lr' in the Table~\ref{tab:hyper}) decayed by half (`lr factor') every 12 epochs (`lr decay int.') for 35 epochs. We use a large and a small PointNet-based~\cite{qi2017pointnet} network (small (`PN-small') and large (`PN-big')) to segment an input TSDF into parts and background. We refer to Tables~\ref{tab:arch_seg_1}, \ref{tab:arch_seg_2} as architectures of `PN-small' and `PN-big' denoted in Table~\ref{tab:hyper}.

\subsection{Fine-tuning on ScanNet data}
To apply to real-world observations, we fine-tune the projections and part segmentation on ScanNet~\cite{dai2017scannet} data using MLCVNet~\cite{xie2020mlcvnet} detections on train scenes. 
We use an Adam optimizer with batch size 64, learning rate 2e-4 (`lr' in the Table~\ref{tab:hyper}) decayed by a factor of 0.2 (`lr factor') every 40 epochs (`lr decay int.') for 80 epochs.

\subsection{Test-time optimization}
For test-time optimization, we optimize for part and shape codes using an Adam optimizer with learning rate of 3e-4 (`lr' in Table~\ref{tab:hyper}) for 500 iterations. The learning rate is multiplied by a factor of 0.1 (`lr factor') after 300 iterations (`lr decay int.')). This part corresponds to the task `Test-time opt.' in Table~\ref{tab:hyper}.

To enable more flexibility to capture input details, we enable optimization of the decoder weights for parts and shape after 400 iterations. We have used the first and the second linear layers of part decoder (`part dec. layers opt.') to optimize simultaneously with latent vectors optimization using Adam optimizer with learning rate of 3e-4 (`lr') for 100 iterations. The learning rate is multiplied by a factor of 0.1 (`lr factor (part dec.)') after 300 iterations (`lr decay int. (part dec.)')). 

In Eqs. (5), (6) we use a weight $w_{trunc}$ for points close to and further away from the surface. 
We have a set $\mathcal{A}_{unif. noise}$ of points that have a distance to surface greater than $d_{trunc}=0.16$m. Having decoded the projection of the shape $\{\tilde{\mathbf{z}}^\textrm{s}\}$, we uniformly sample  points around decoded shape no closer than 0.2m to the surface of this shape, and assign truncation distance $d_{trunc}$ to these points. We also add them to the set $\mathcal{A}_{unif. noise}$. For the shape decoder and for the set $\mathcal{A}_{unif. noise}$ we set $w_{trunc} = 5.0$ (`$w_{trunc}$ (shape unif. noise)' in the Table~\ref{tab:hyper}); for part decoder we set $w_{trunc} = 20.0$ (`$w_{trunc}$ (parts unif. noise)'). Additionally, while optimizing the particular part $k$ during test-time optimization we also use the points corresponding to other parts as noise with distance $d_{trunc}$ and denote this set of points as $\mathcal{A}_{part noise}$. Adding this set into optimization is necessary to decrease intersections between different part geometries after test-time optimization. We set $w_{trunc} = 5.0$ (`$w_{trunc}$ (part noise)') for this set of points. Finally, in Eq. (4) we use an additional weight for loss consistency term, for which we set $w_{cons} = 200.0$.

To encourage geometric completeness during test-time optimization, we sample points with distances to surface from the decoded shape $\mathcal{S}$ and parts $\{\mathcal{P}_k\}$ (decoded from $\{\tilde{\mathbf{z}}^\textrm{s}\}$ and $\{\tilde{\mathbf{z}}_k^\textrm{p}\}$), and add them to TSDF $D$ (`add pts. to shape' in the Table~\ref{tab:hyper}) or $\{D^{\textrm{p}}\}_{\textrm{p}=1}^{N_{parts}}$ (`add pts. to parts') to the regions where points in $\mathcal{S}$ or $\{\mathcal{P}_k\}$ are present and non-background points with distance $d < d_{trunc}$ in $D$ and $\{D^{\textrm{p}}\}_{\textrm{p}=1}^{N_{parts}}$ (which we call \textit{meaningful} points) are missing. We add only those points from $\mathcal{S}$ and $\{\mathcal{P}_k\}$ which are not closer than $d_{thr}$ to meaningful points.

Finally, we scale the coordinates of input TSDF with a scale factor (`scale factor') to align better to the learned canonical space of synthetic shapes.

Optimization for each part takes approximately 25 seconds.

\subsection{Hyperparameters search} 

We determined hyperparameters for training and TTO on a hold-out validation set. There are parameters that affect training and TTO more than others, such as the number of decoder layers to use parameters from for TTO for both shape and part decoders and learning rate for both training and TTO. Parameters of decoder layers enable more flexibility in optimization, but decrease implicit regularization from learned part priors, resulting in less geometry consistency. The proper choice of the learning rate for TTO is important for accurate geometry reconstruction and avoiding unstable optimization. All weights for the loss components in TTO have a wide range of appropriate parameters ($\pm20$ for $w_{trunc}$ and $\pm50$ for $w_{cons}$).

\begin{table*}[tp]
\centering
\resizebox{0.8\textwidth}{!}{
\begin{tabular}{c|c|c|c|c|c|c|c}
\toprule
    Task & Parameter & Chair & Table & Cabinet & Bookshelf & Bed & Trashcan \\
\midrule
    Train decoder (shape) & \# epochs & 2000 & 2400 & 8000 & 8000 & 16000 & 16000 \\
    Train decoder (shape) & batch size & 24 & 24 & 24 & 24 & 24 & 24 \\
    Train decoder (shape) & optimizer & Adam & Adam & Adam & Adam & Adam & Adam \\
    Train decoder (shape) & lr (weights) & 5e-5 & 1e-4 & 1e-4 & 1e-4 & 1e-4 & 1e-4 \\
    Train decoder (shape) & lr factor (weights) & 0.5 & 0.5 & 0.5 & 0.5 & 0.5 & 0.5 \\
    Train decoder (shape) & lr decay int. (weights) & 500 & 600 & 1500 & 1500 & 4000 & 3000 \\
    Train decoder (shape) & lr (lat.) & 1e-4 & 2e-4 & 2e-4 & 2e-4 & 2e-4 & 2e-4 \\
    Train decoder (shape) & lr factor (lat.) & 0.5 & 0.5 & 0.5 & 0.5 & 0.5 & 0.5 \\
    Train decoder (shape) & lr decay int. (lat.) & 500 & 600 & 1500 & 1500 & 4000 & 3000 \\
\midrule
    Train decoder (parts) & \# epochs & 1100 & 1400 & 2000 & 2000 & 10000 & 10000 \\
    Train decoder (parts) & batch size & 48 & 48 & 48 & 48 & 48 & 48 \\
    Train decoder (parts) & optimizer & Adam & Adam & Adam & Adam & Adam & Adam \\
    Train decoder (parts) & lr (weights) & 5e-5 & 1e-4 & 1e-4 & 1e-4 & 1e-4 & 1e-4 \\
    Train decoder (parts) & lr factor (weights) & 0.5 & 0.5 & 0.5 & 0.5 & 0.5 & 0.5 \\
    Train decoder (parts) & lr decay int. (weights) & 400 & 400 & 800 & 800 & 3600 & 3600 \\
    Train decoder (parts) & lr (lat.) & 1e-4 & 2e-4 & 2e-4 & 2e-4 & 2e-4 & 2e-4 \\
    Train decoder (parts) & lr factor (lat.) & 0.5 & 0.5 & 0.5 & 0.5 & 0.5 & 0.5 \\
    Train decoder (parts) & lr decay int. (lat.) & 400 & 400 & 800 & 800 & 3600 & 3600 \\
\midrule
    Train projection & \# epochs & 35 & 30 & 60 & 60 & 250 & 200 \\
    Train projection & batch size & 64 & 64 & 64 & 64 & 64 & 64 \\
    Train projection & optimizer & Adam & Adam & Adam & Adam & Adam & Adam \\
    Train projection & lr & 1e-3 & 1e-3 & 1e-3 & 1e-3 & 1e-3 & 1e-3 \\
    Train projection & lr factor & 0.5 & 0.5 & 0.5 & 0.5 & 0.5 & 0.5 \\
    Train projection & lr decay int. & 12 & 20 & 40 & 40 & 150 & 120 \\
    Train projection & segm. network & PN-small & PN-big & PN-big & PN-big & PN-big & PN-big \\
\midrule
    Fine-tune & \# epochs & 80 & 80 & 120 & 120 & 125 & 70 \\
    Fine-tune & batch size & 64 & 64 & 64 & 64 & 64 & 64 \\
    Fine-tune & optimizer & Adam & Adam & Adam & Adam & Adam & Adam \\
    Fine-tune & lr & 2e-4 & 2e-4 & 2e-4 & 2e-4 & 2e-4 & 2e-4 \\
    Fine-tune & lr factor & 0.2 & 0.2 & 0.2 & 0.2 & 0.2 & 0.2 \\
    Fine-tune & lr decay int. & 40 & 40 & 60 & 60 & 60 & 40 \\
\midrule
    Test-time opt. & \# iterations & 500 & 500 & 500 & 500 & 500 & 500 \\
    Test-time opt. & optimizer & Adam & Adam & Adam & Adam & Adam & Adam \\
    Test-time opt. & lr (lat.) & 3e-4 & 3e-4 & 3e-4 & 3e-4 & 3e-4 & 3e-4 \\
    Test-time opt. & lr factor (lat.) & 0.1 & 0.1 & 0.1 & 0.1 & 0.1 & 0.1 \\
    Test-time opt. & lr decay int. (lat.) & 300 & 300 & 300 & 300 & 300 & 300 \\
    Test-time opt. & shape dec. layers opt. & - & - & - & - & 1,2 & 1,2 \\
    Test-time opt. & lr (shape dec.) & - & - & - & - & 1e-4 & 1e-4 \\
    Test-time opt. & lr factor (shape dec.) & - & - & - & - & 0.1 & 0.1 \\
    Test-time opt. & lr decay int. (shape dec.) & - & - & - & - & 300 & 300 \\
    Test-time opt. & part dec. layers opt. & 1,2 & 1,2 & 1,2 & 1,2 & - & - \\
    Test-time opt. & lr (part dec.) & 3e-4 & 3e-4 & 3e-4 & 3e-4 & - & - \\
    Test-time opt. & lr factor (part dec.) & 0.1 & 0.1 & 0.1 & 0.1 & - & - \\
    Test-time opt. & lr decay int. (part dec.) & 300 & 300 & 300 & 300 & - & - \\
    Test-time opt. & $w_{trunc}$ (shape unif. noise) & 5.0 & 3.0 & 3.0 & 3.0 & 1.0 & 1.0 \\
    Test-time opt. & $w_{trunc}$ (parts unif. noise) & 20.0 & 12.0 & 12.0 & 12.0 & 1.0 & 5.0 \\
    Test-time opt. & $w_{trunc}$ (part noise) & 5.0 & 3.0 & 10.0 & 10.0 & 10.0 & 5.0 \\
    Test-time opt. & $w_{cons}$ & 200.0 & 300.0 & 300.0 & 300.0 & 30.0 & 200.0 \\
    Test-time opt. & add pts. to shape & \checkmark & \checkmark & \checkmark & \checkmark & \checkmark & - \\
    Test-time opt. & dist thr. (shape) & 0.16m & 0.16m & 0.5m & 0.5m & 0.75m & - \\
    % Test-time opt. & CD thr. (shape) & 90.0 & 90.0 & 90.0 & 90.0 & 50.0 & - \\
    Test-time opt. & add pts. to parts & - & \checkmark & \checkmark & \checkmark & \checkmark & - \\
    Test-time opt. & dist thr. (part) & - & 0.16m & 0.5m & 0.5m & 0.75m & - \\
    % Test-time opt. & CD thr. (part) & - & 90.0 & 90.0 & 90.0 & 50.0 & - \\
    Test-time opt. & scale factor & 1.2 & 1.2 & 1.1 & 1.1 & 1.2 & 1.2 \\
\bottomrule
\end{tabular}
}
\caption{Hyperparameters used for training submodels used in our framework.}
\label{tab:hyper}
\end{table*}

\subsection{Network architecture}

We also provide extensive information about the architecture of every submodel that we use in our framework. Table~\ref{tab:arch_encoder} shows the architecture of voxel encoder that we use to encode an input occupancy grid. Table~\ref{tab:arch_child_decoder} shows the architecture of a module that predicts the part decomposition of an input object. The architectures of a small PointNet-like~\cite{qi2017pointnet} network and a big PointNet-like network that we use to segment an input TSDF are shown in Tables~\ref{tab:arch_seg_1}, \ref{tab:arch_seg_2}. Finally, we provide details about the architecture of shape and parts MLP decoders in Tables~\ref{tab:arch_shape_sdf}, \ref{tab:arch_parts_sdf}.

\begin{table*}[tp]
\centering
\resizebox{0.8\textwidth}{!}{
\begin{tabular}{c|c|c|c|c|c|c|c}
\toprule
    Encoder & Input Layer & Type & Input Size & Output Size & Kernel Size & Stride & Padding \\
\midrule
    conv0 & scan occ. grid & Conv3D & (1, 32, 32, 32) & (32, 16, 16, 16) & (5, 5, 5) & (2, 2, 2) & (2, 2, 2) \\
    gnorm0 & conv0 & GroupNorm & (32, 16, 16, 16) & (32, 16, 16, 16) & - & - & - \\
    relu0 & gnorm0 & ReLU & (32, 16, 16, 16) & (32, 16, 16, 16) & - & - & - \\
    pool1 & relu0 & MaxPooling & (32, 16, 16, 16) & (32, 8, 8, 8) & (2, 2, 2) & (2, 2, 2) & (0, 0, 0) \\
    conv1 & pool1 & Conv3D & (32, 8, 8, 8) & (64, 8, 8, 8) & (3, 3, 3) & (1, 1, 1) & (1, 1, 1) \\
    gnorm1 & conv1 & GroupNorm & (64, 8, 8, 8) & (64, 8, 8, 8) & - & - & - \\
    relu1 & gnorm1 & ReLU & (64, 8, 8, 8) & (64, 8, 8, 8) & - & - & - \\
    pool2 & relu1 & MaxPooling & (64, 8, 8, 8) & (64, 4, 4, 4) & (2, 2, 2) & (2, 2, 2) & (0, 0, 0) \\
    conv2 & pool2 & Conv3D & (64, 4, 4, 4) & (128, 2, 2, 2) & (5, 5, 5) & (2, 2, 2) & (2, 2, 2) \\
    gnorm2 & conv2 & GroupNorm & (128, 2, 2, 2) & (128, 2, 2, 2) & - & - & - \\
    relu2 & gnorm2 & ReLU & (128, 2, 2, 2) & (128, 2, 2, 2) & - & - & - \\
    pool3 & relu2 & MaxPooling & (128, 2, 2, 2) & (128, 1, 1, 1) & (2, 2, 2) & (2, 2, 2) & (0, 0, 0) \\
    conv3 & pool3 & Conv3D & (128, 1, 1, 1) & (256, 1, 1, 1) & (3, 3, 3) & (1, 1, 1) & (1, 1, 1) \\
    gnorm3 & conv3 & GroupNorm & (256, 1, 1, 1) & (256, 1, 1, 1) & - & - & - \\
    relu3 & gnorm3 & ReLU & (256, 1, 1, 1) & (256, 1, 1, 1) & - & - & - \\
    shape feature & relu3 & Flatten & (256, 1, 1, 1) & (256) & - & - & - \\
\bottomrule
\end{tabular}
}
\caption{Layer specification for detected object encoder.}
\label{tab:arch_encoder}
\end{table*}

\begin{table*}[tp]
\centering
\resizebox{0.8\textwidth}{!}{
\begin{tabular}{c|c|c|c|c}
\toprule
    Child decoder & Input Layer & Type & Input Size & Output Size \\
\midrule
    lin\_proj & shape feature & ReLU(Linear) & 256 & 256 \\
    node feature & lin\_proj & ReLU(Linear) & 256 & 256 \\
\midrule
    lin0 & node feature & Linear & 256 & 2560 \\
    relu0 & lin0 & ReLU & 2560 & 2560 \\
    reshape0 & relu0 & Reshape & 2560 & (10, 256) \\
    node\_exist & reshape0 & Linear & (10, 256) & (10, 1) \\
\midrule
    concat0 & (reshape0, reshape0) & Concat. & (10, 256), (10, 256) & (10, 10, 512) \\
    lin1 & concat0 & Linear & (10, 10, 512) & (10, 10, 256) \\
    relu1 & lin1 & ReLU & (10, 10, 256) & (10, 10, 256) \\
    edge\_exist & relu1 & Linear & (10, 10, 256) & (10, 10, 1) \\
\midrule
    mp & (relu1, edge\_exist, reshape0) & Mes. Passing & (10, 10, 256), (10, 10, 1), (10, 256) & (10, 768) \\
    lin2 & mp & Linear & (10, 768) & (10, 256) \\
    relu2 & lin2 & ReLU & (10, 256) & (10, 256) \\
    node\_sem & relu2 & Linear & (10, 256) & (10, \#classes) \\
\midrule
    lin3 & relu2 & Linear & (10, 256) & (10, 256) \\
    (10, child feature)  & lin3 & ReLU & (10, 256) & (10, 256) \\
\midrule
    lin4 & node feature & ReLU(Linear) & 256 & 256 \\
    rotation\_cls & lin3 & Linear & 256 & 12 \\
\bottomrule
\end{tabular}
}
\caption{Layer specification for decoding an object into its semantic part structure.}
\label{tab:arch_child_decoder}
\end{table*}

\begin{table*}[tp]
\centering
\resizebox{0.8\textwidth}{!}{
\begin{tabular}{c|c|c|c|c}
\toprule
    Pts. classifier (small) & Input Layer & Type & Input Size & Output Size \\
\midrule
    input feature & (TSDF, node feature, rotation\_cls) & Concat. & (\#pts, 4), 256, 12 & (\#pts, 272) \\
    lin\_cls\_0 & input feature & ReLU(Linear) & (\#pts, 272) & (\#pts, 128) \\
    lin\_cls\_1 & lin\_cls\_0 & ReLU(Linear) & (\#pts, 128) & (\#pts, 128) \\
    lin\_cls\_2 & lin\_cls\_1 & ReLU(Linear) & (\#pts, 128) & (\#pts, 128) \\
    glob\_feat\_0 & lin\_cls\_2 & MaxPooling1D & (\#pts, 128) & (1, 128) \\
    glob\_feat\_1 & glob\_feat\_0 & Repeat & (1, 128) & (\#pts, 128) \\
    lin\_cls\_3 & (lin\_cls\_1, glob\_feat\_1) & Concat & (\#pts, 128), (\#pts, 128) & (\#pts, 256) \\
    lin\_cls\_4 & lin\_cls\_3 & ReLU(Linear) & (\#pts, 256) & (\#pts, 128) \\
    lin\_cls\_5 & lin\_cls\_4 & ReLU(Linear) & (\#pts, 128) & (\#pts, 128) \\
    lin\_cls\_6 & lin\_cls\_5 & Linear & (\#pts, 128) & (\#pts, \#classes) \\
\bottomrule
\end{tabular}
}
\caption{Layer specification for segmenting input TSDF using small PointNet-like network.}
\label{tab:arch_seg_1}
\end{table*}

\begin{table*}[tp]
\centering
\resizebox{0.8\textwidth}{!}{
\begin{tabular}{c|c|c|c|c}
\toprule
    Pts. classifier (big) & Input Layer & Type & Input Size & Output Size \\
\midrule
    input feature & (TSDF, node feature, rotation\_cls) & Concat. & (\#pts, 4), 256, 12 & (\#pts, 272) \\
    lin\_cls\_0 & input feature & ReLU(Linear) & (\#pts, 272) & (\#pts, 256) \\
    lin\_cls\_1 & lin\_cls\_0 & ReLU(Linear) & (\#pts, 256) & (\#pts, 128) \\
    lin\_cls\_2 & lin\_cls\_1 & ReLU(Linear) & (\#pts, 128) & (\#pts, 128) \\
    glob\_feat\_0 & lin\_cls\_2 & MaxPooling1D & (\#pts, 128) & (1, 128) \\
    glob\_feat\_1 & glob\_feat\_0 & Repeat & (1, 128) & (\#pts, 128) \\
    lin\_cls\_3 & lin\_cls\_2 & ReLU(Linear) & (\#pts, 128) & (\#pts, 64) \\
    lin\_cls\_4 & lin\_cls\_3 & ReLU(Linear) & (\#pts, 64) & (\#pts, 64) \\
    glob\_feat\_2 & lin\_cls\_4 & MaxPooling1D & (\#pts, 64) & (1, 64) \\
    glob\_feat\_3 & glob\_feat\_2 & Repeat & (1, 64) & (\#pts, 64) \\
    lin\_cls\_5 & (lin\_cls\_1, glob\_feat\_1, glob\_feat\_3) & Concat & (\#pts, 128), (\#pts, 128), (\#pts, 64) & (\#pts, 320) \\
    lin\_cls\_6 & lin\_cls\_5 & ReLU(Linear) & (\#pts, 320) & (\#pts, 128) \\
    lin\_cls\_7 & lin\_cls\_6 & ReLU(Linear) & (\#pts, 128) & (\#pts, 64) \\
    lin\_cls\_8 & lin\_cls\_7 & Linear & (\#pts, 64) & (\#pts, \#classes) \\
\bottomrule
\end{tabular}
}
\caption{Layer specification for segmenting input TSDF using big PointNet-like network.}
\label{tab:arch_seg_2}
\end{table*}

\begin{table*}[tp]
\centering
\resizebox{0.8\textwidth}{!}{
\begin{tabular}{c|c|c|c|c}
\toprule
    Implicit decoder & Input Layer & Type & Input Size & Output Size \\
\midrule
    lin\_proj\_0 & node feature & ReLU(Linear) & 256 & 512 \\
    lin\_proj\_1 & lin\_proj\_0 & ReLU(Linear) & 512 & 512 \\
    lin\_proj\_2 & lin\_proj\_1 & ReLU(Linear) & 512 & 512 \\
    lin\_proj\_3 & lin\_proj\_2 & ReLU(Linear) & 512 & 512 \\
    lin\_proj\_4 & lin\_proj\_3 & Linear & 512 & 256 \\
    lin\_pts\_0 & (lin\_proj\_4, TSDF pts.) & Concat. & 256, 63 & 319 \\
    
    lin\_pts\_1 & lin\_pts\_0 & Linear & 319 & 512 \\
    lin\_bn\_1 & lin\_pts\_1 & BatchNorm & 512 & 512 \\
    lin\_relu\_1 & lin\_bn\_1 & ReLU & 512 & 512 \\
    lin\_drop\_1 & lin\_relu\_1 & Dropout & 512 & 512 \\
    
    lin\_pts\_2 & lin\_pts\_1 & Linear & 512 & 512 \\
    lin\_bn\_2 & lin\_pts\_2 & BatchNorm & 512 & 512 \\
    lin\_relu\_2 & lin\_bn\_2 & ReLU & 512 & 512 \\
    lin\_drop\_2 & lin\_relu\_2 & Dropout & 512 & 512 \\
    
    lin\_pts\_3 & lin\_pts\_2 & Linear & 512 & 512 \\
    lin\_bn\_3 & lin\_pts\_3 & BatchNorm & 512 & 512 \\
    lin\_relu\_3 & lin\_bn\_3 & ReLU & 512 & 512 \\
    lin\_drop\_3 & lin\_relu\_3 & Dropout & 512 & 512 \\
    
    lin\_pts\_4 & lin\_pts\_3 & Linear & 512 & 512 - dim(lin\_pts\_0) \\
    lin\_bn\_4 & lin\_pts\_4 & BatchNorm & 512 - dim(lin\_pts\_0) & 512 - dim(lin\_pts\_0) \\
    lin\_relu\_4 & lin\_bn\_4 & ReLU & 512 - dim(lin\_pts\_0) & 512 - dim(lin\_pts\_0) \\
    lin\_drop\_4 & lin\_relu\_4 & Dropout & 512 - dim(lin\_pts\_0) & 512 - dim(lin\_pts\_0) \\
    
    lin\_pts\_5 & (lin\_pts\_0, lin\_drop\_4) & Concat. & dim(lin\_pts\_0), 512 - dim(lin\_pts\_0) & 512 \\
    lin\_bn\_5 & lin\_pts\_5 & BatchNorm & 512 & 512 \\
    lin\_relu\_5 & lin\_bn\_5 & ReLU & 512 & 512 \\
    lin\_drop\_5 & lin\_relu\_5 & Dropout & 512 & 512 \\
    
    lin\_pts\_6 & lin\_pts\_5 & Linear & 512 & 512 \\
    lin\_bn\_6 & lin\_pts\_6 & BatchNorm & 512 & 512 \\
    lin\_relu\_6 & lin\_bn\_6 & ReLU & 512 & 512 \\
    lin\_drop\_6 & lin\_relu\_6 & Dropout & 512 & 512 \\
    
    lin\_pts\_7 & lin\_pts\_6 & Linear & 512 & 512 \\
    lin\_bn\_7 & lin\_pts\_7 & BatchNorm & 512 & 512 \\
    lin\_relu\_7 & lin\_bn\_7 & ReLU & 512 & 512 \\
    lin\_drop\_7 & lin\_relu\_7 & Dropout & 512 & 512 \\
    
    lin\_pts\_8 & lin\_pts\_7 & Linear & 512 & 1 \\
    lin\_tanh\_7 & lin\_pts\_8 & Tanh & 1 & 1 \\
\bottomrule
\end{tabular}
}
\caption{Layer specification for implicit shape decoder.}
\label{tab:arch_shape_sdf}
\end{table*}

\begin{table*}[tp]
\centering
\resizebox{0.8\textwidth}{!}{
\begin{tabular}{c|c|c|c|c}
\toprule
    Implicit decoder & Input Layer & Type & Input Size & Output Size \\
\midrule
    lin\_proj\_0 & child feature & ReLU(Linear) & 256 & 512 \\
    lin\_proj\_1 & lin\_proj\_0 & ReLU(Linear) & 512 & 512 \\
    lin\_proj\_2 & lin\_proj\_1 & ReLU(Linear) & 512 & 512 \\
    lin\_proj\_3 & lin\_proj\_2 & ReLU(Linear) & 512 & 512 \\
    lin\_proj\_4 & lin\_proj\_3 & Linear & 512 & 256 \\
    lin\_pts\_0 & (lin\_proj\_4, part cls. one-hot, TSDF pts.) & Concat. & 256, \#parts, 63 & 319 + \#parts \\
    
    lin\_pts\_1 & lin\_pts\_0 & Linear & 319 + \#parts & 512 \\
    lin\_bn\_1 & lin\_pts\_1 & BatchNorm & 512 & 512 \\
    lin\_relu\_1 & lin\_bn\_1 & ReLU & 512 & 512 \\
    lin\_drop\_1 & lin\_relu\_1 & Dropout & 512 & 512 \\
    
    lin\_pts\_2 & lin\_pts\_1 & Linear & 512 & 512 \\
    lin\_bn\_2 & lin\_pts\_2 & BatchNorm & 512 & 512 \\
    lin\_relu\_2 & lin\_bn\_2 & ReLU & 512 & 512 \\
    lin\_drop\_2 & lin\_relu\_2 & Dropout & 512 & 512 \\
    
    lin\_pts\_3 & lin\_pts\_2 & Linear & 512 & 512 \\
    lin\_bn\_3 & lin\_pts\_3 & BatchNorm & 512 & 512 \\
    lin\_relu\_3 & lin\_bn\_3 & ReLU & 512 & 512 \\
    lin\_drop\_3 & lin\_relu\_3 & Dropout & 512 & 512 \\
    
    lin\_pts\_4 & lin\_pts\_3 & Linear & 512 & 512 - dim(lin\_pts\_0) \\
    lin\_bn\_4 & lin\_pts\_4 & BatchNorm & 512 - dim(lin\_pts\_0) & 512 - dim(lin\_pts\_0) \\
    lin\_relu\_4 & lin\_bn\_4 & ReLU & 512 - dim(lin\_pts\_0) & 512 - dim(lin\_pts\_0) \\
    lin\_drop\_4 & lin\_relu\_4 & Dropout & 512 - dim(lin\_pts\_0) & 512 - dim(lin\_pts\_0) \\
    
    lin\_pts\_5 & (lin\_pts\_0, lin\_drop\_4) & Concat. & dim(lin\_pts\_0), 512 - dim(lin\_pts\_0) & 512 \\
    lin\_bn\_5 & lin\_pts\_5 & BatchNorm & 512 & 512 \\
    lin\_relu\_5 & lin\_bn\_5 & ReLU & 512 & 512 \\
    lin\_drop\_5 & lin\_relu\_5 & Dropout & 512 & 512 \\
    
    lin\_pts\_6 & lin\_pts\_5 & Linear & 512 & 512 \\
    lin\_bn\_6 & lin\_pts\_6 & BatchNorm & 512 & 512 \\
    lin\_relu\_6 & lin\_bn\_6 & ReLU & 512 & 512 \\
    lin\_drop\_6 & lin\_relu\_6 & Dropout & 512 & 512 \\
    
    lin\_pts\_7 & lin\_pts\_6 & Linear & 512 & 512 \\
    lin\_bn\_7 & lin\_pts\_7 & BatchNorm & 512 & 512 \\
    lin\_relu\_7 & lin\_bn\_7 & ReLU & 512 & 512 \\
    lin\_drop\_7 & lin\_relu\_7 & Dropout & 512 & 512 \\
    
    lin\_pts\_8 & lin\_pts\_7 & Linear & 512 & 1 \\
    lin\_tanh\_7 & lin\_pts\_8 & Tanh & 1 & 1 \\
\bottomrule
\end{tabular}
}
\caption{Layer specification for implicit part decoder.}
\label{tab:arch_parts_sdf}
\end{table*}

\end{document}